\documentclass{ieeeaccess}

\usepackage[cmex10]{amsmath}
\usepackage[utf8]{inputenc}
\usepackage[T1]{fontenc}
\usepackage{bm}
\usepackage{graphicx}
\usepackage{float}
\usepackage{subcaption}
\usepackage[linesnumbered,ruled,vlined,onelanguage]{algorithm2e}
\usepackage{algorithmic}
\usepackage{mathtools}
\usepackage{array}

\usepackage{array}
\usepackage{makecell}
\usepackage{csvsimple}
\usepackage{booktabs}
\usepackage{xargs}
\usepackage{color, colortbl}
\usepackage{footmisc}

\usepackage{ifthen}

\def\BibTeX{{\rm B\kern-.05em{\sc i\kern-.025em b}\kern-.08em
		T\kern-.1667em\lower.7ex\hbox{E}\kern-.125emX}}

\newcommand\orcidicon[1]{\href{https://orcid.org/#1}{\makebox[0.2cm]{
			\includegraphics[width=0.2cm]{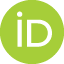}
}}}

\usepackage{xargs}

\usepackage{fourier} 
\usepackage{array}
\usepackage{makecell}
\usepackage{csvsimple}
\usepackage{booktabs}

\renewcommand{\theenumii}{\theenumi.\arabic{enumii}.}
\usepackage{tabularx}
\usepackage{caption,setspace}
\usepackage{xargs}

\usepackage{hyperref} %<--- Load after everything else

\begin{document}
	\history{Date of publication 03 August 2023, date of current version 25 August 2023.}
	\doi{10.1109/ACCESS.2023.3301727}
	
	\title{MeROS: SysML-based Metamodel for ROS-based Systems}
	
	\author{\uppercase{Tomasz Winiarski}\authorrefmark{\orcidicon{0000-0002-9316-3284}1},
		\IEEEmembership{Member, IEEE}}
	\address[1]{Warsaw University of Technology, Institute of Control and Computation Engineering, Nowowiejska 15/19, 00-665 Warsaw, Poland}
	\tfootnote{Research was funded by the Centre for Priority Research Area Artificial Intelligence and Robotics of Warsaw University of Technology within the Excellence Initiative: Research University (IDUB) programme}
	
	\markboth
	{T. Winiarski \headeretal: MeROS: SysML-based Metamodel for ROS-based Systems}
	{T. Winiarski \headeretal: MeROS: SysML-based Metamodel for ROS-based Systems}
	
	\corresp{Corresponding author: Tomasz Winiarski (e-mail: tomasz.winiarski@pw.edu.pl).}
	
	\begin{abstract}
		The complexity of today's robot control systems implies difficulty in developing them efficiently and reliably. Systems engineering (SE) and frameworks come to help. The framework metamodels are needed to support the standardisation and correctness of the created application models. Although the use of frameworks is widespread nowadays, for the most popular of them, Robot Operating System (ROS), a~contemporary  metamodel has been missing so far. This article proposes a~new metamodel for ROS called MeROS, which addresses the running system and developer workspace. The ROS comes in two versions: ROS~1 and ROS~2. The metamodel includes both versions. In particular, the latest ROS~1 concepts are considered, such as nodelet, action, and metapackage. An essential addition to the original ROS concepts is the grouping of these concepts, which provides an opportunity to illustrate the system's decomposition and varying degrees of detail in its presentation. The metamodel is derived from the requirements and verified on the practical example of Rico assistive robot. The matter is described in a standardised way in SysML (Systems Modeling Language). Hence, common development tools that support SysML can help develop robot controllers in the spirit of SE.
	\end{abstract}
	
	\begin{keywords}
		Systems engineering, Robot Operating System ROS, ROS~1, ROS~2, Robotic framework, SysML, Platform Specific Model PSM
	\end{keywords}
	
	\titlepgskip=-15pt
	
	\maketitle
	
\section{Introduction}
\label{sec:intro}	
	The development of civilisation has led to an increase in the importance of robotics. Many modern robotic systems are complex. To create them as effectively and reliably as possible, it is necessary to follow systems engineering (SE), where metamodels play an essential role~\cite{bezivin2004search,schmidt2006model,kent2002model}.
	Robots, especially complex ones, are mostly controlled with usage of software. Hence, in robotics, SE is inextricably linked with software engineering, where frameworks have been crucial for many years \cite{mnkandla2009software,shehory2014agent}.
	Diverse robotics frameworks have been developed so far \cite{inigo2012robotics,tsardoulias2017robotic,hentout2016survey}. Some steps towards standardisation have been made in recent years, and ROS (Robot Operating System) has come to the fore. Stand-alone ROS~1 (ROS version~1)~\cite{quigley2009ros} is unsuitable for hard RT (Real Time) systems, so one of the solutions in practical applications (e.g., \cite{lages2014architecture,buys2011haptic,pages2016tiago,Seredynski-fabric-romoco-2019,kornuta-bpan-2020,cholewinski2015software}), is to integrate ROS~1 with Orocos \cite{bruyninckx2001open,bruyninckx2002orocos}. Over time, ROS~1 has evolved to, among other things, improve its performance. Known and crucial problems in the face of some contemporary applications (e.g. cybersecurity, RT performance) led to the development of a~new version of the framework, ROS~2 \cite{maruyama2016exploring,park2020real}.   

	ROS~1 has evolved considerably from the initial distributions. Its metamodels created so far are now incomplete and outdated (sec.~\ref{sec:related-work}). According to ROS metrics\footnote{\url{https://discourse.ros.org/t/2022-ros-metrics-report/29594}}, new ROS~1 distributions have practically replaced older ones in terms of distro downloads stats. Above indications point to the need to formulate an up-to-date, recent metamodel for the latest versions of ROS~1 and ROS~2, which this article undertakes by presenting the new metamodel for ROS -- MeROS.
	
	The robotic models can be subdivided~\cite{de2021survey} into Platform Independent Models (PIM), e.g., \cite{zielinski2017variable,zielinski2010motion,tasker2020,earl2020}, and Platform Specific Models (PSM). The metamodels of ROS, including MeROS, belong to PSM and should answer to the component nature of ROS \cite{Figat:2022:RAS,wenger2016model}.
		
 MeROS is founded on SysML (Systems Modeling Language) \cite{omg-sysml16,Friedenthal:2015}, a profile of UML (Unified Modeling Language). Modelling in languages from the UML family addresses a number of important aspects of systems engineering \cite{chaudron2012effective}. These include the use cases [UCX]:
	\begin{itemize}
		\item $[$UC1] Systems' documentation and presentation,
		\item $[$UC2] Effective analysis of systems, especially in interdisciplinary teams (graphical language is more understandable for non-specialists in the field), 
		\item $[$UC3] Defects detection,
		\item $[$UC4] Integration of new collaborators into the development team,
		\item $[$UC5] Resuming work after a break,
		\item $[$UC6] Extension and modification of existing systems,
		\item $[$UC7] Support the implementation of new systems,
		\item $[$UC8] Migration of systems.   
	\end{itemize}
	
	In practice, documentation is created both prior to implementation and, in many cases, through a process of reverse  engineering \cite{canfora2007new} for existing systems. Agile-type strategies involve modifying the documentation as the project develops \cite{habib2021systematic}.
			
	The following presentation starts with formulating the requirements (sec.~\ref{sec:requirements}) for the MeROS metamodel. These requirements are allocated to the metamodel that is described in sec.~\ref{sec:metamodel}. The way to present a~model of a~specific application based on MeROS is presented on a practical example in sec.~\ref{sec:application}.
	A comparison of the MeROS with similar metamodels is presented in sec.~\ref{sec:related-work}. The paper is finalised with conclusions (sec.~\ref{sec:conslusions}).

\section{Metamodel requirements}
\label{sec:requirements}
	The requirements [RX] formulation process for MeROS metamodel is multi-stage and iterative. In the beginning, the initial requirements were formulated based on: (i) literature review (both scientific and ROS wiki/community sources), (ii) author experience from supervising and supporting ROS-based projects, and finally, (iii) author experience from EARL (Embodied Agent-based cybeR-physical control systems modelling Language) \cite{earl2020} PIM development and its applications (e.g. \cite{tasker2020,karwowski2021hubero,en14206693-grav-comp}). Verification of draft versions of MeROS by its practical applications led to an iterative reformulation of requirements and MeROS itself. The article presents the final version of both MeROS metamodel and the requirements it originates from.
	
\begin{figure}[htb]
		\centering
		\begin{center}
			{\includegraphics[scale=0.7]{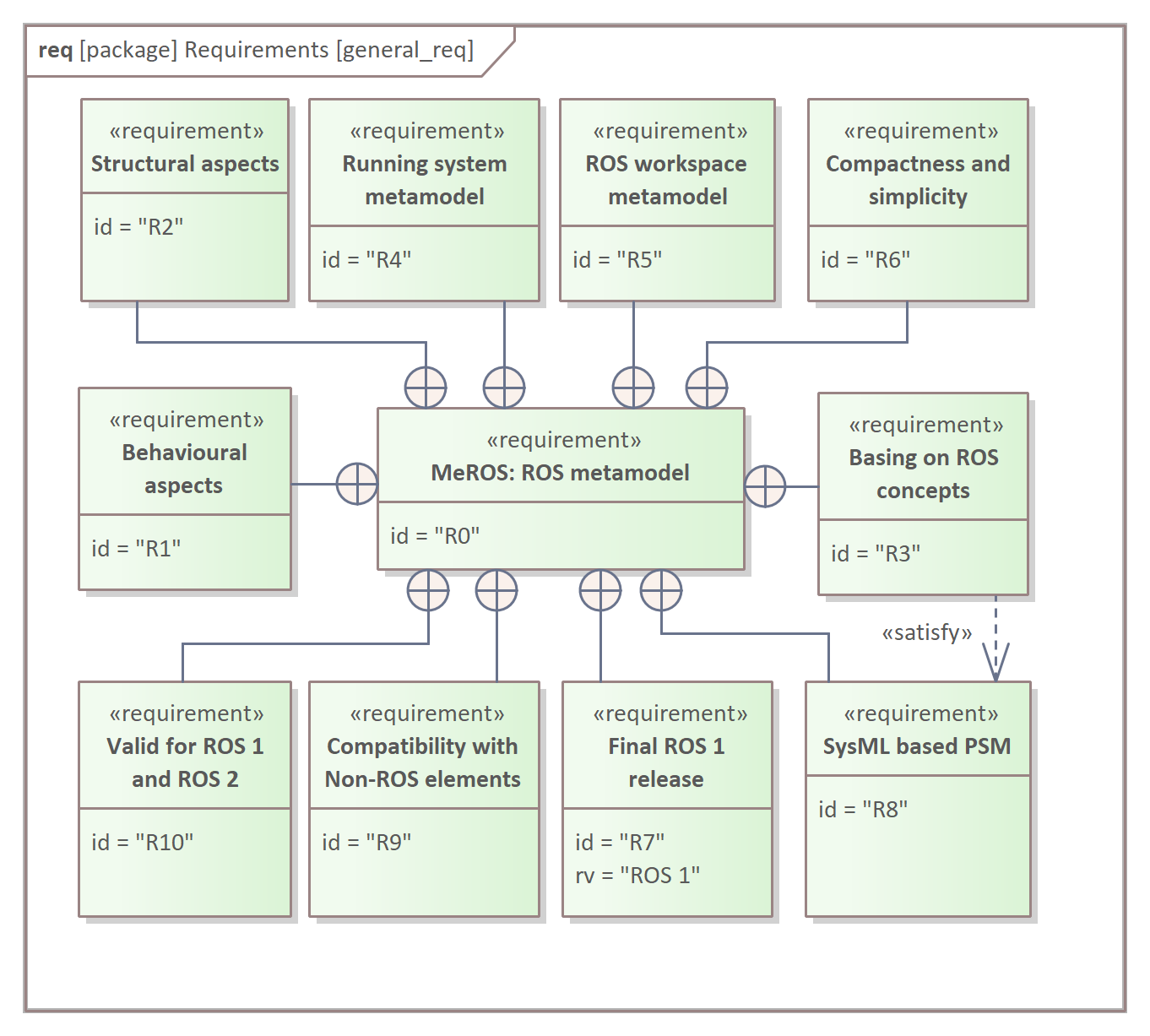}}
		\end{center}
		\caption{General requirements.} 
		\label{fig:general_req}
	\end{figure}

	MeROS requirements are depicted on a~number of dedicated SysML diagrams. The requirements are organised in a~tree-like nesting structure, with additional internal relations, and labelled following this structure. The general requirements are presented in Fig.~\ref{fig:general_req}. Here, and in the following diagrams, the elements (components, relations) specific for a particular version of ROS (ROS~1 or ROS~2) are labelled with an ,,rv''  tag with the ROS version that the element is specific for. The lack of a tag means the element is general for both ROS versions.
	
% DLA SKLADU ZMIENIONO WIELKOSC Z 0.7	

	The SysML models have two main parts: behavioural [R1] and structural [R2]. MeROS aims to cover ROS concepts [R3] and not change their labels as long as possible, to maintain conformity and intuitiveness. The ROS system is two-faced. While it is executed [R4], it has a~specific structure and behaviour, but from the developers' point of view, the workspace [R5] is the exposed aspect. The model should be compact and straightforward [R6] rather than unnecessarily elaborate and complicated. One of the assumptions that stand out MeROS from other ROS metamodels is conformity with the final ROS~1 release [R7] (Noetic Ninjemys). Although the SysML-based MeROS is classified into PSM [R8], it should be compatible with Non-ROS elements [R9]. Finally, MeROS metamodel should be valid both for ROS~1 and ROS~2.
	
\begin{figure}[htb]
		\centering
		\begin{center}
			{\includegraphics[scale=0.7]{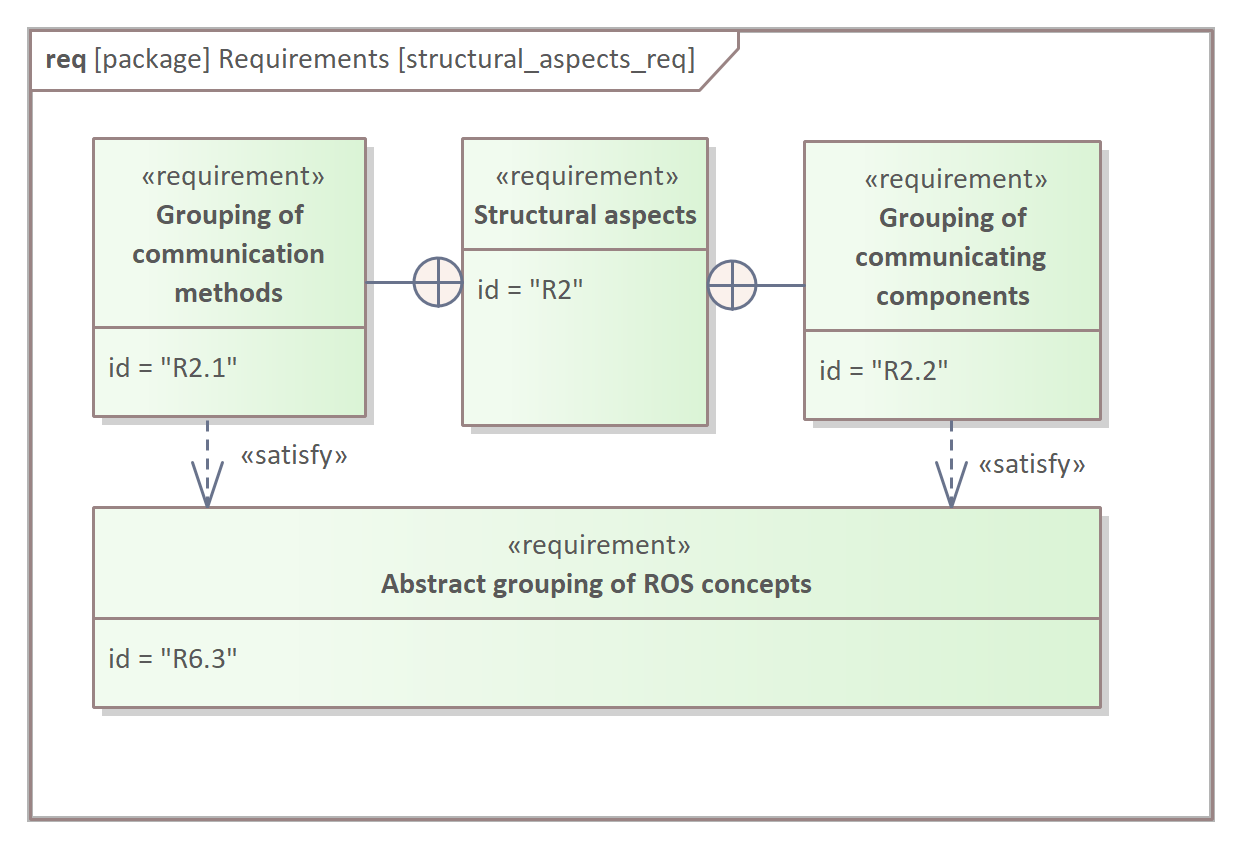}}
		\end{center}
		\caption{Structural aspects requirements.} 
		\label{fig:structural_aspects_req}
	\end{figure}

	The system's structural aspects requirements are presented in Fig.~\ref{fig:structural_aspects_req}.
	
\begin{figure}[htb]
		\centering
		\begin{center}
			{\includegraphics[scale=0.7]{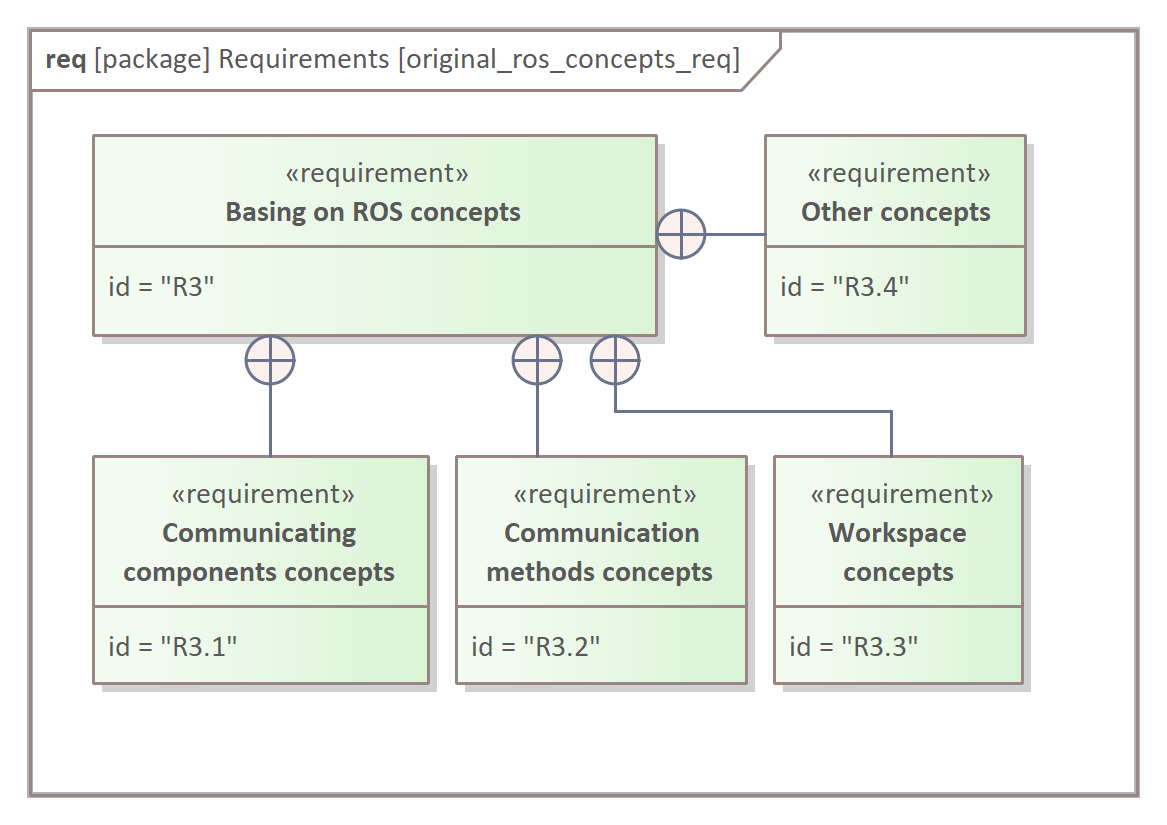}}
		\end{center}
		\caption{ROS concepts requirements.} 
		\label{fig:ros_concepts_req}
	\end{figure}

	A~vital addition to the original ROS concepts is the abstract grouping of: (i) communicating methods [R2.1] and (ii) communicating components [R2.2]. The motivation for the introduction of these aggregates is presented further on. It should be noted that several ROS concepts group other concepts in a particular way, especially to deploy the system. Action aggregates Topics and Services (in ROS~2), ROS~1 Node aggregates Nodelets, and ROS~2 Component Container aggregates Nodes.
	The ROS concepts that MeROS models are organised into four major classes (Fig.~\ref{fig:ros_concepts_req}): (i) Communicating components [R3.1], (ii) Communication methods [R3.2], (iii) Workspace [R3.3], and (iv) Other [R3.4].
	
% DLA SKLADU ZMIENIONO WIELKOSC Z 0.7
	
	Communicating components [R3.1] are: (i) ROS Node, (ii) ROS Nodelet, (iii) ROS plugin, and (iv) ROS library. Both plugin and library let to share the same code between various Nodes or ROS~1 specific Nodelets. Two ROS nodes are mandatory to execute the ROS~1 system: (i) ROS Master and (ii) rosout.
%% DLA SKLADU ZMIENIONO WIELKOSC Z 0.7
%	\begin{figure}[htb]
%		\centering
%		\begin{center}
%			{\includegraphics[scale=0.66]{img/requirement_pkg/communicating_components_req.png}}
%		\end{center}
%		\caption{Communicating components requirements.} 
%		\label{fig:communicating_components_req}
%	\end{figure}
%	Communication methods are depicted in Fig.~\ref{fig:communication_concepts_req}.
	 Three methods of communication are considered [R3.2] with their inter-component connections and data structures: (i) ROS Topic, its Message and connection, (ii) ROS Service comprising data structure and connection, and finally (iii) ROS Action including data structure and connection.
	% DLA SKLADU ZMIENIONO WIELKOSC Z 0.7
%	\begin{figure}[htb]
%		\centering
%		\begin{center}
%			{\includegraphics[scale=0.66]{img/requirement_pkg/communication_concepts_req.png}}
%		\end{center}
%		\caption{Communication concepts requirements.} 
%		\label{fig:communication_concepts_req}
%	\end{figure}
	Workspace concept [R3.3] comprises: (i) ROS Package [R3.3.1] and (ii) Metapackage [R3.3.2] introduced in the latest releases of ROS~1.
%	% DLA SKLADU ZMIENIONO WIELKOSC Z 0.7
%	\begin{figure}[htb]
%		\centering
%		\begin{center}
%			{\includegraphics[scale=0.66]{img/requirement_pkg/workspace_concepts_req.png}}
%		\end{center}
%		\caption{Workspace concepts requirements.} 
%		\label{fig:workspace_concepts_req}
%	\end{figure}
	Other concepts [R3.4] include four elements: (i) ROS Parameter Server manages (ii) ROS Parameters, (iii) roscore forms a collection of programs and nodes that are pre-requisites of a ROS~1-based system. Finally, (iv) ROS Namespace reflects the ROS concept to organise nodes and communication connections. Both ROS Master and rosout are executed with roscore. ROS Parameter Server is a part of ROS Master.
%	% DLA SKLADU ZMIENIONO WIELKOSC Z 0.7
%	\begin{figure}[htb]
	%		\centering
	%		\begin{center}
		%			{\includegraphics[scale=0.66]{img/requirement_pkg/other_concepts_req.png}}
		%		\end{center}
	%		\caption{Other ROS concepts requirements.} 
	%		\label{fig:other_concepts_req}
	%	\end{figure}

%	Fig.~\ref{fig:roscore_req} presents additional roscore-related relations.

%	\begin{figure}[htb]
%		\centering
%		\begin{center}
%			{\includegraphics[scale=0.7]{img/requirement_pkg/roscore_req.png}}
%		\end{center}
%		\caption{roscore related requirements dependences.} 
%		\label{fig:roscore_req}
%	\end{figure}

\begin{figure}[htb]
		\centering
		\begin{center}
			{\includegraphics[scale=0.69]{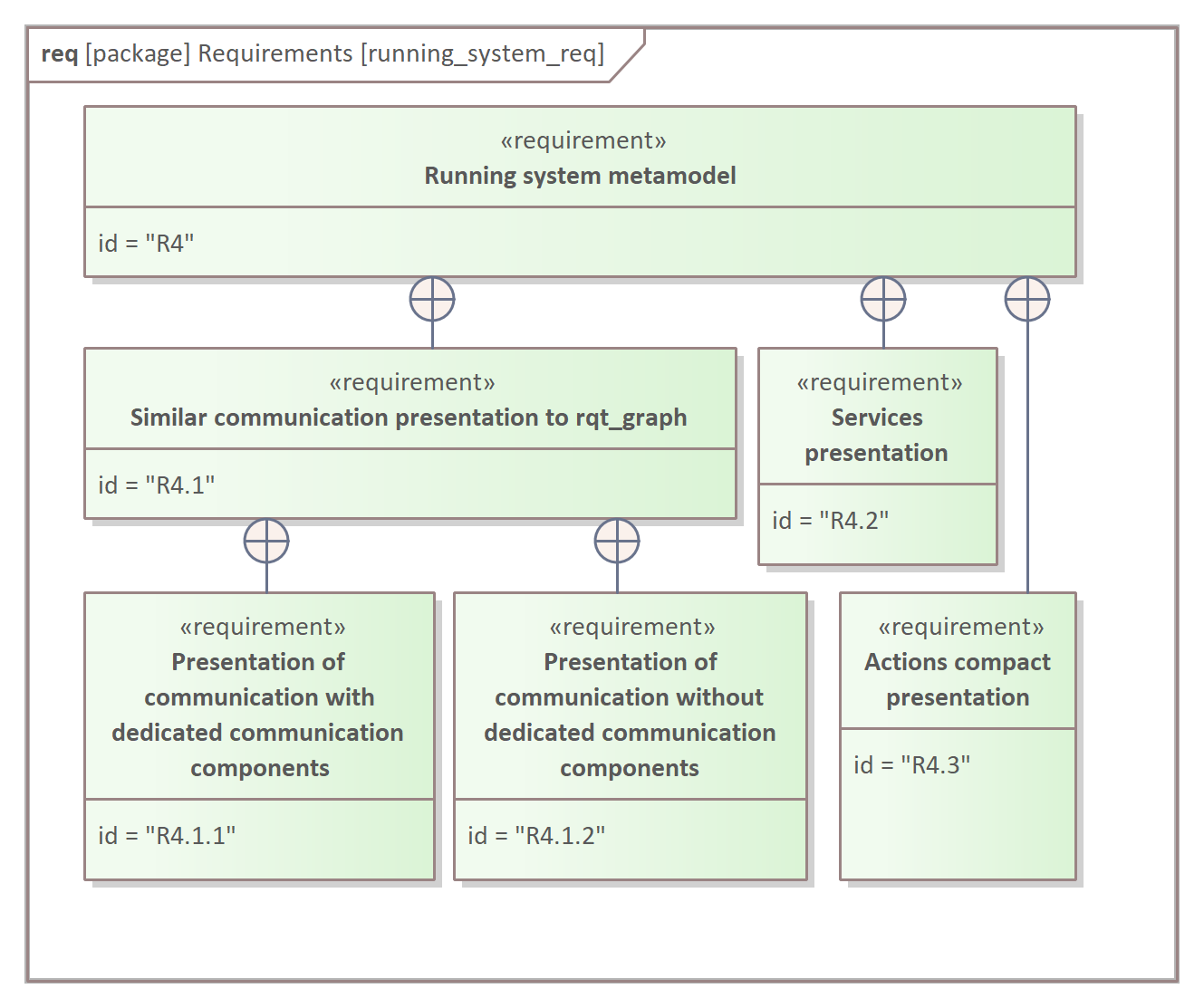}}
		\end{center}
		\caption{Running system requirements.} 
		\label{fig:running_system_req}
	\end{figure}

	To achieve intuitiveness, MeROS presents a Running system structure (Fig.~\ref{fig:running_system_req}) following ROS rqt\_graph pattern [R4.1]. In particular, there are two ways to visualise communication, including [R4.1.1] and without [R4.1.2]
	dedicated communication components.
	 The dedicated components are especially useful in the presentation when many communication components use the same topic both on the publisher and the subscriber side. In opposition, the expression of topic names on arrows connecting communicating components, i.e., without dedicated communication components, let to reduce the number of components needed to depict communication for many topics and a~low number of communicating components. The other advantage of using dedicated communication components is that the particular connection can be split into several diagrams (e.g. ibd (internal block diagram) or sd (sequence diagram)), where the same object represents this connection in every associated diagram. Services [R4.2] and actions [R4.3] should be depicted as an addition to the presentation of the particular topics. It should be noted that rqt\_graph represents actions as a~number of topics and services. In MeROS, the topics and services being part of an action can be aggregated, which reduces the number of depicted connections.
		
\begin{figure}[htb]
		\centering
		\begin{center}
			{\includegraphics[scale=0.69]{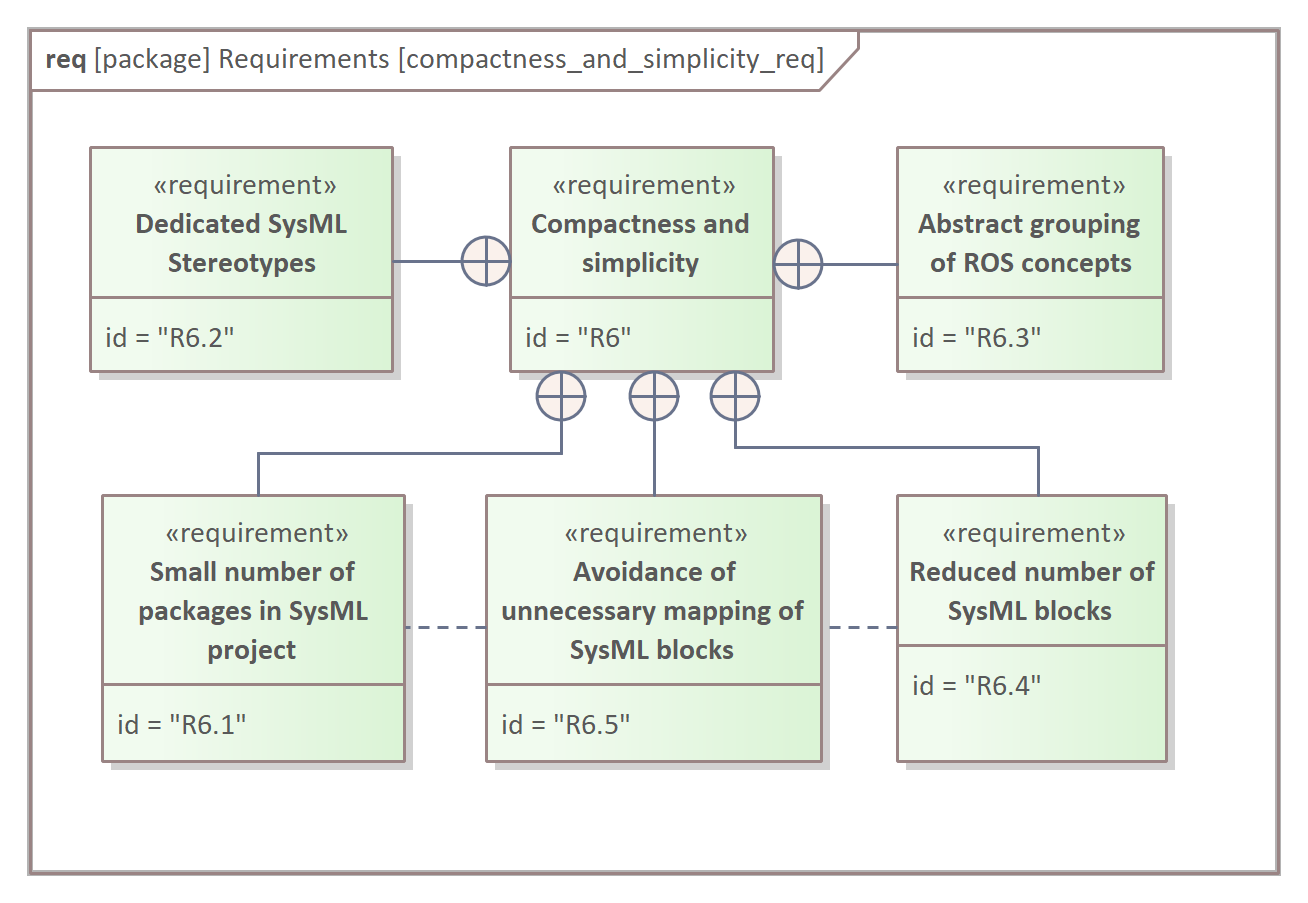}}
		\end{center}
		\caption{Compactness and simplicity requirements.} 
		\label{fig:compactness_and_simplicity_req}
	\end{figure}

	The compactness and simplicity [R6] and its nesting requirements are presented in Fig.~\ref{fig:compactness_and_simplicity_req}. 
	
	A SysML project to develop and represent MeROS metamodel should consist of a~small number of packages [R6.1], but still, the packages should distinguish the major aspects of development process: (i) metamodel requirements formulation, (ii) metamodel itself, and (iii) metamodel realizations/applications.
	Dedicated SysML stereotypes [R6.2] are introduced to MeROS to replace the direct block specialization representation on diagrams and improve the legibility and compactness of diagrams.
	The grouping of concepts [R6.3] has diverse aims. It enables the presentation of the system part in a~general, PIM-like abstract way, on the logical level rather than a~detailed, PSM-like implementation one. The aggregation reduces the number of objects represented on the diagram to highlight the essential aspects and stay compact and consistent in presentation.
	The number of SysML blocks should be reduced to a reasonable level [R6.4]. Both [R6.1] and [R6.4] help in the Avoidance of unnecessary mapping of SysML blocks [R6.5].

\begin{figure}[htb]
		\centering
		\begin{center}
			{\includegraphics[scale=0.7]{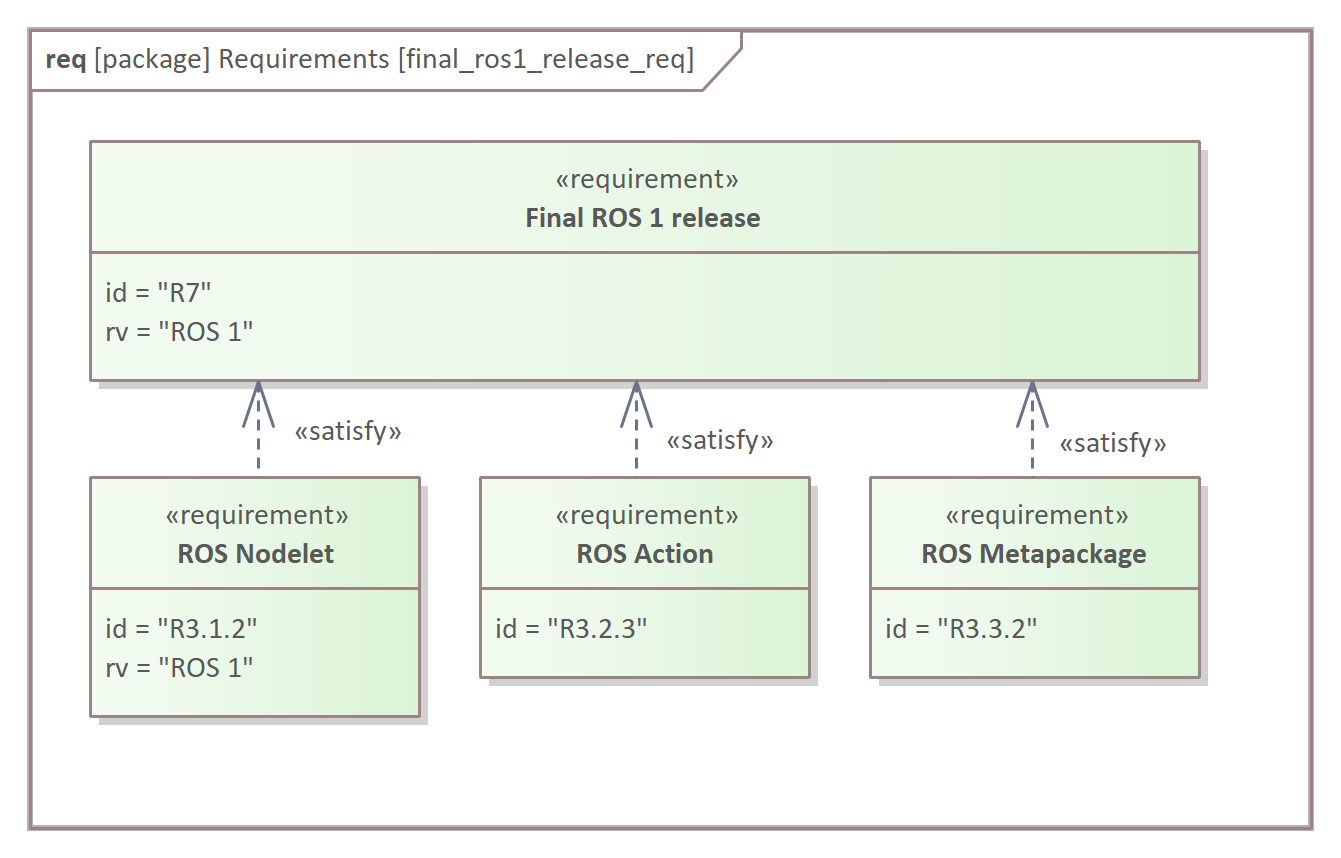}}
		\end{center}
		\caption{Final ROS 1 release requirements.} 
		\label{fig:final_ros1_release_req}
	\end{figure}

	There are three elements in the requirements set that satisfy the evolution of the ROS~1  finalised with its ultimate release -- Noetic Ninjemys -- (Fig.~\ref{fig:final_ros1_release_req}): (i) ROS Nodelet  (introduced primarily to increase the efficiency of ROS components switching), (ii) ROS Action , and (iii) ROS Metapackage.
	
\section{MeROS metamodel}
\label{sec:metamodel}
	
	MeROS metamodel is formulated according to the requirements discussed in the previous section. Sec.~\ref{sec:metamodel-composition} presents MeROS blocks' structural composition, and sec.~\ref{sec:metamodel-communication} describes inter-component communication. From the metamodel perspective, the structural aspects [R2] are formulated in both sections, while behavioural [R1] is in the latter. The diagrams comprise selected requirements being allocated to expose the MeROS metamodel development process. 
	
\begin{figure}[H]
		\centering
		\begin{center}
			{\includegraphics[scale=0.7]{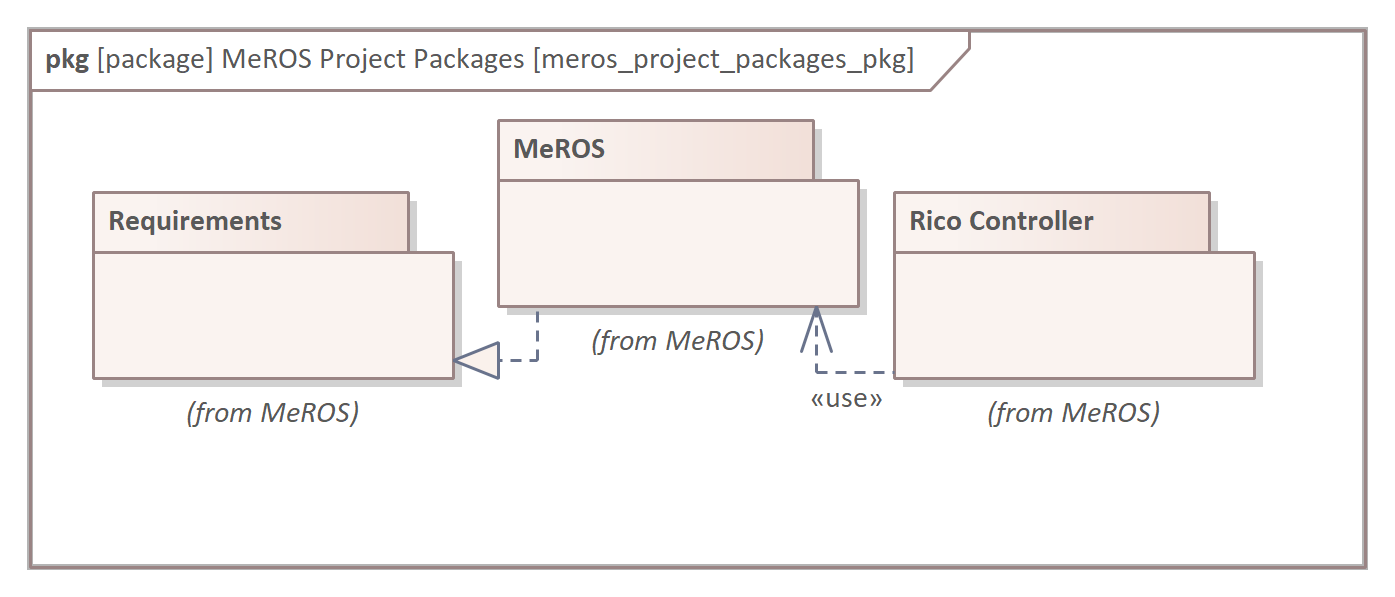}}
		\end{center}
		\caption{MeROS project SysML packages, where Rico Controller is an exemplary realisation of MeROS metamodel.} 
		\label{fig:meros_project_packages_pkg}
	\end{figure}

	The MeROS diagrams were created in the Enterprise Architect development tool within the SysML project [R8] and organised in three packages [R6.1] (Fig.~\ref{fig:meros_project_packages_pkg}): (i) Requirement Model related to requirements formulation and analysis, (ii) MeROS -- the metamodel itself, (iii) Rico Controller -- the exemplary ROS~1 application of MeROS described in sec.~\ref{sec:application-example}. The stereotypes are introduced in MeROS metamodel with the dedicated MeROS profile [R6.2].
	
		% DLA SKLADU ZMIENIONO WIELKOSC Z 0.7
	
\subsection{Metamodel composition}
\label{sec:metamodel-composition}

	The degree of specificity of a metamodel is a compromise between its comprehensiveness (and, therefore, more general formulation) and a more accurate representation of a particular subclass of specific implementations. The metamodel contains compositions of elements and other primary relationships. Attributes and operations range widely, in particular between ROS~1 and ROS~2. Hence, their inclusion would lead to overgrowth and complication of the metamodel [R6]. Models derived from the metamodel can define their operations and new relations specific to a particular system. 
	
\begin{figure}[htb]
		\centering
		\begin{center}
			{\includegraphics[scale=0.7]{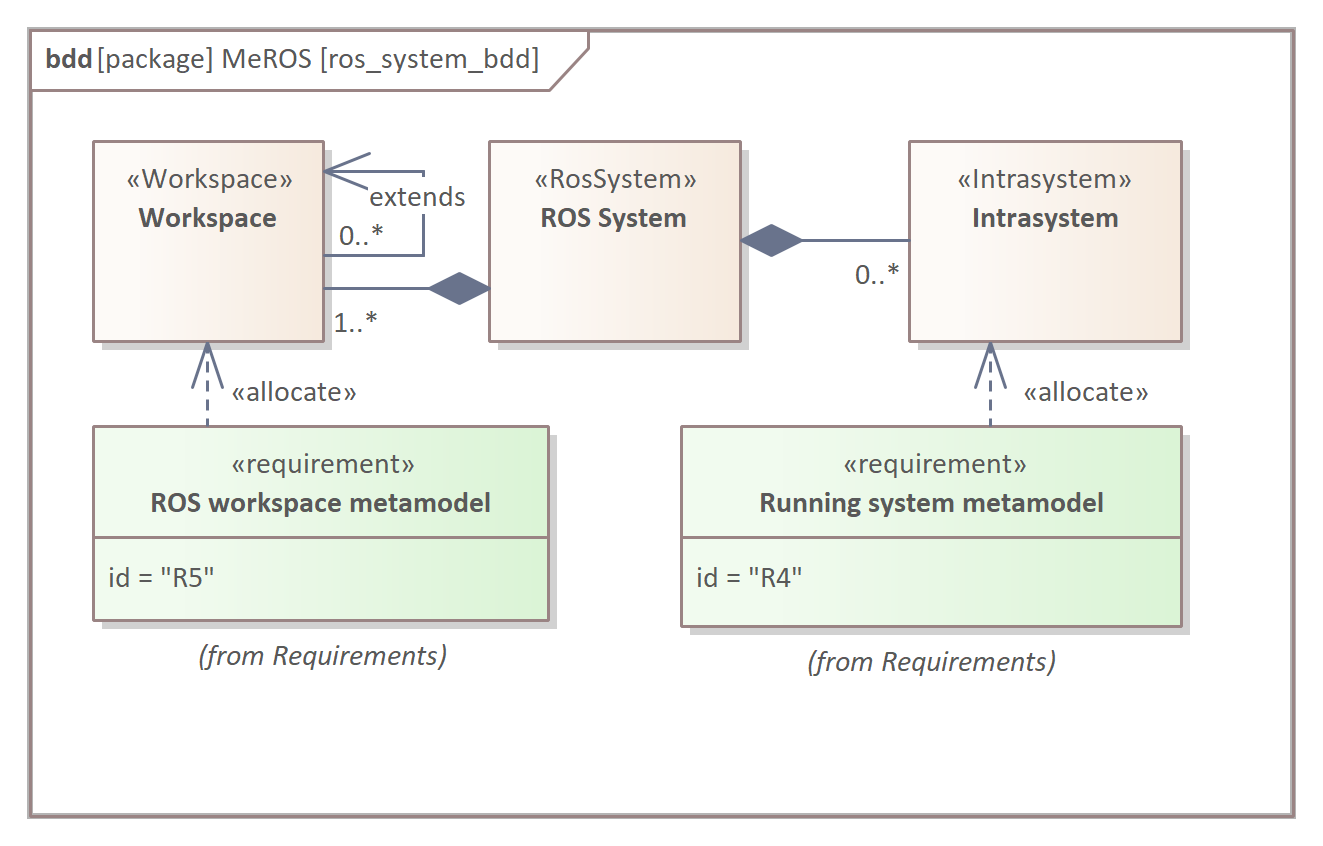}}
		\end{center}
		\caption{ROS system general composition -- bdd.} 
		\label{fig:ros_system_bdd}
	\end{figure}

	The SysML blocks reflect ROS concepts [R3], and their composition is depicted in bdd (block definition diagrams). The metamodel is formulated in a~single SysML package. Hence, Workspaces and Intrasystems are composed into ROS System (Fig.~\ref{fig:ros_system_bdd}). 
	
		% DLA SKLADU ZMIENIONO WIELKOSC Z 0.7
	
	Consequently, some concepts (e.g., Node) occur in Workspaces and Intrasystems. It reduces the number of SysML blocks in the metamodel [R6.4] and eliminates the need for unnecessary mapping of SysML blocks [R6.5].
		
\begin{figure}[htb]
		\centering
		\begin{center}
			{\includegraphics[scale=0.7]{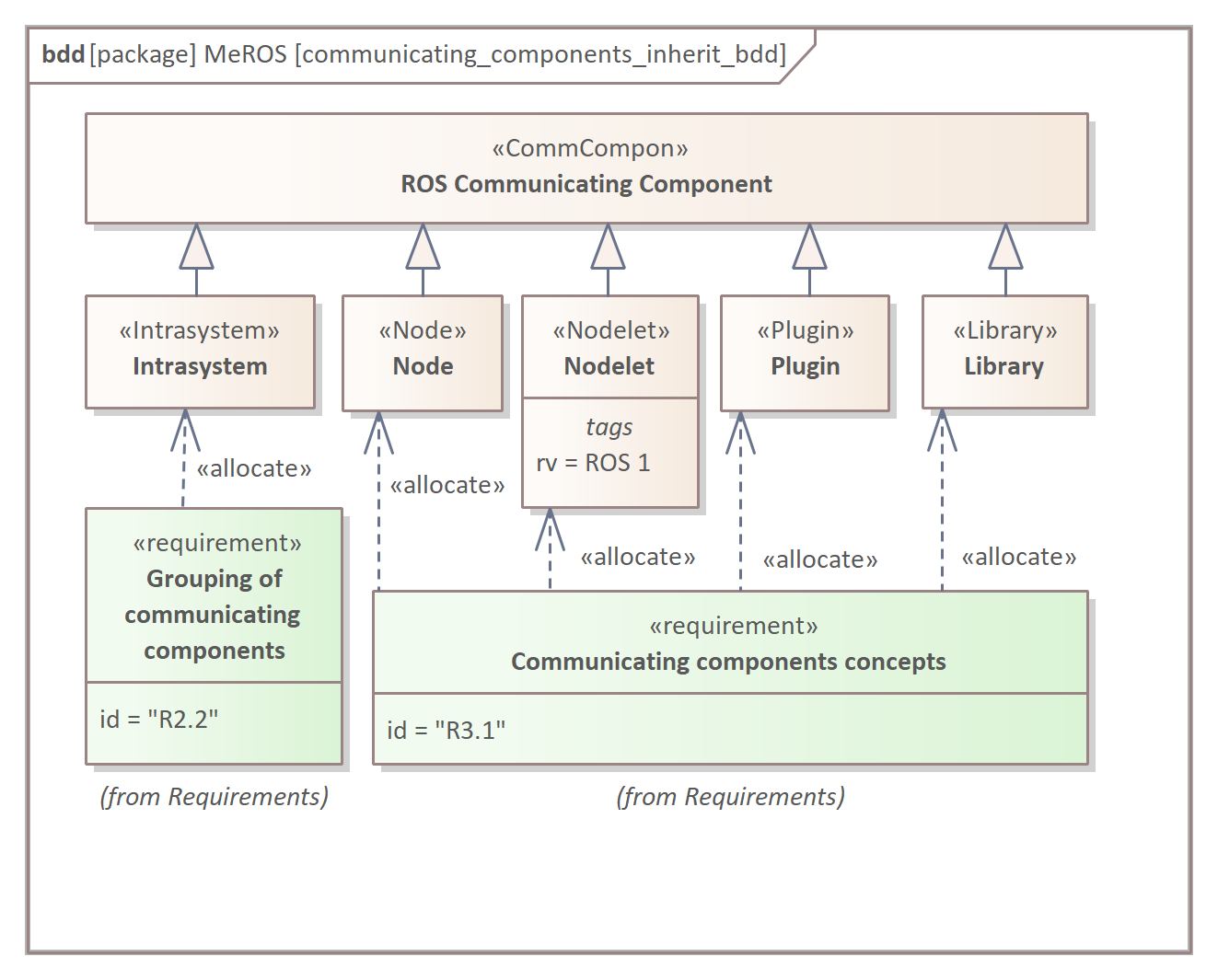}}
		\end{center}
		\caption{Communicating Component and specialised blocks -- bdd.} 
		\label{fig:communicating_components_bdd}
	\end{figure}

	In MeROS, a~Communicating Component (Fig.~\ref{fig:communicating_components_bdd}) is a~crucial abstraction of a~number of ROS concepts to represent their standardised role regarding communication. 
	
		% DLA SKLADU ZMIENIONO WIELKOSC Z 0.7
	
\begin{figure}[htb]
		\centering
		\begin{center}
			{\includegraphics[scale=0.7]{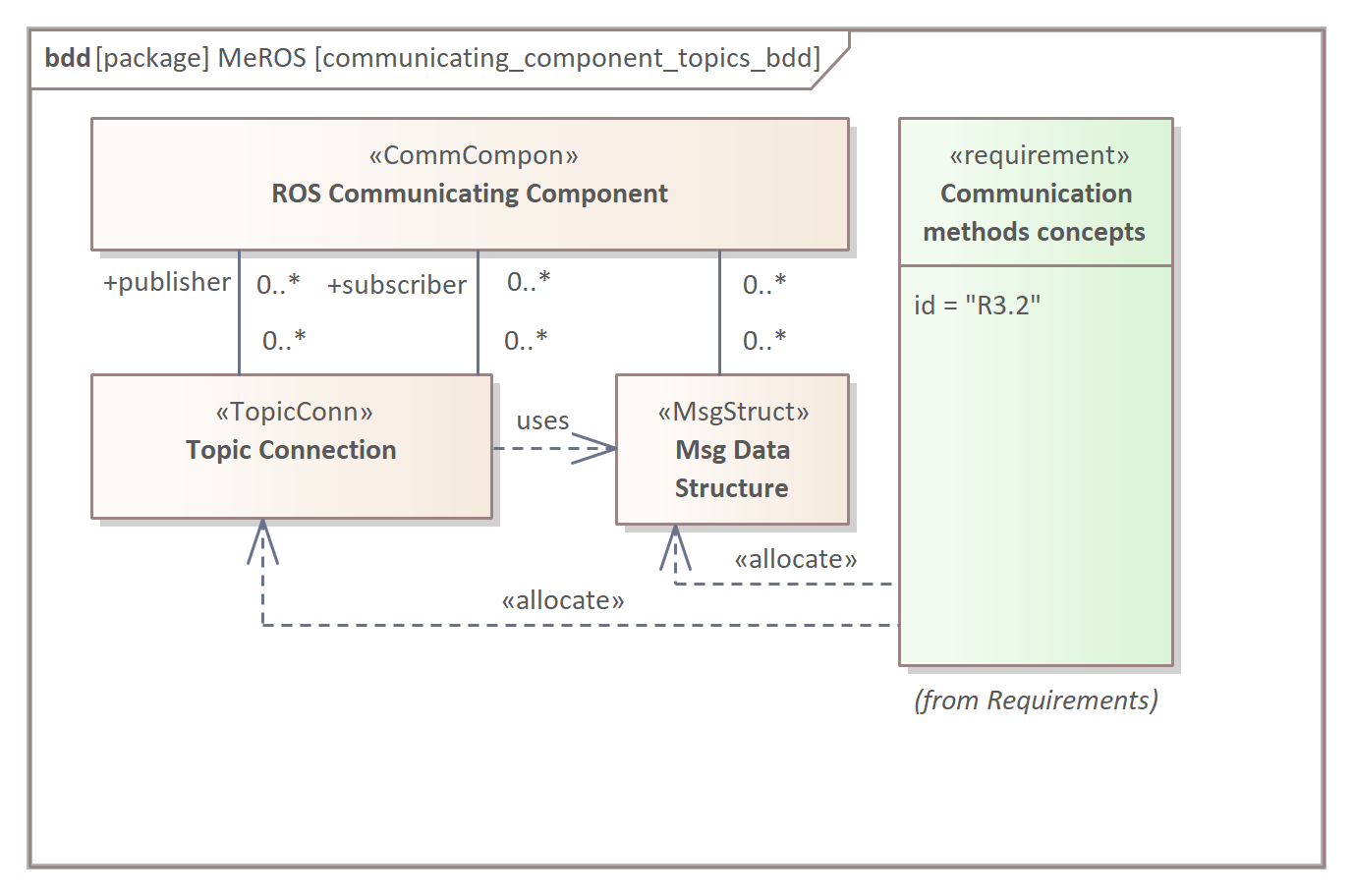}}
		\end{center}
		\caption{Communicating Component relations -- topics -- bdd.} 
		\label{fig:communicating_component_topics_bdd}
	\end{figure}

	It should be noted that behavioural aspects of a~particular model specified in MeROS can be formulated by operation specification as an act (activity), sd (sequence), or stm (state machine) diagrams. The Intrasystem is one of the aggregates added to the base ROS concepts in MeROS. 
	For clarity, relations of Communicating Components are depicted in several diagrams. Fig.~\ref{fig:communicating_component_topics_bdd} considers Topics and their Data Structures. Here, the Communicating Component can act as a publisher or a subscriber.	
	 
	% DLA SKLADU ZMIENIONO WIELKOSC Z 0.7
	
\begin{figure}[htb]
		\centering
		\begin{center}
			{\includegraphics[scale=0.69]{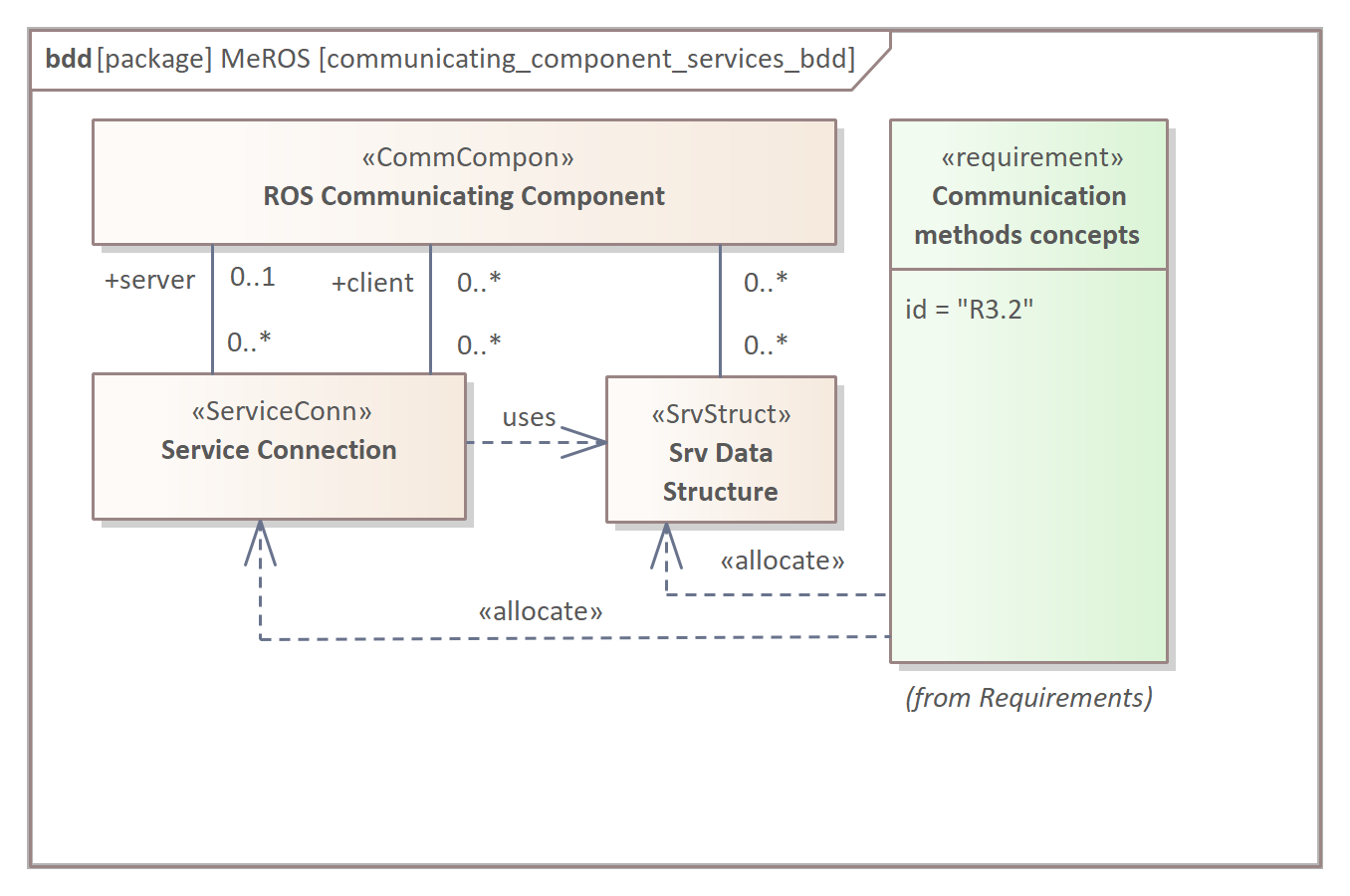}}
		\end{center}
		\caption{Communicating Component relations -- services -- bdd.} 
		\label{fig:communication_blocks_services_bdd}
	\end{figure}

	Fig.~\ref{fig:communication_blocks_services_bdd} depicts Services and their Data Structures. In this case, the Communicating Component can act as a server or a client.	
	
		% DLA SKLADU ZMIENIONO WIELKOSC Z 0.7
	
\begin{figure}[htb]
		\centering
		\begin{center}
			{\includegraphics[scale=0.68]{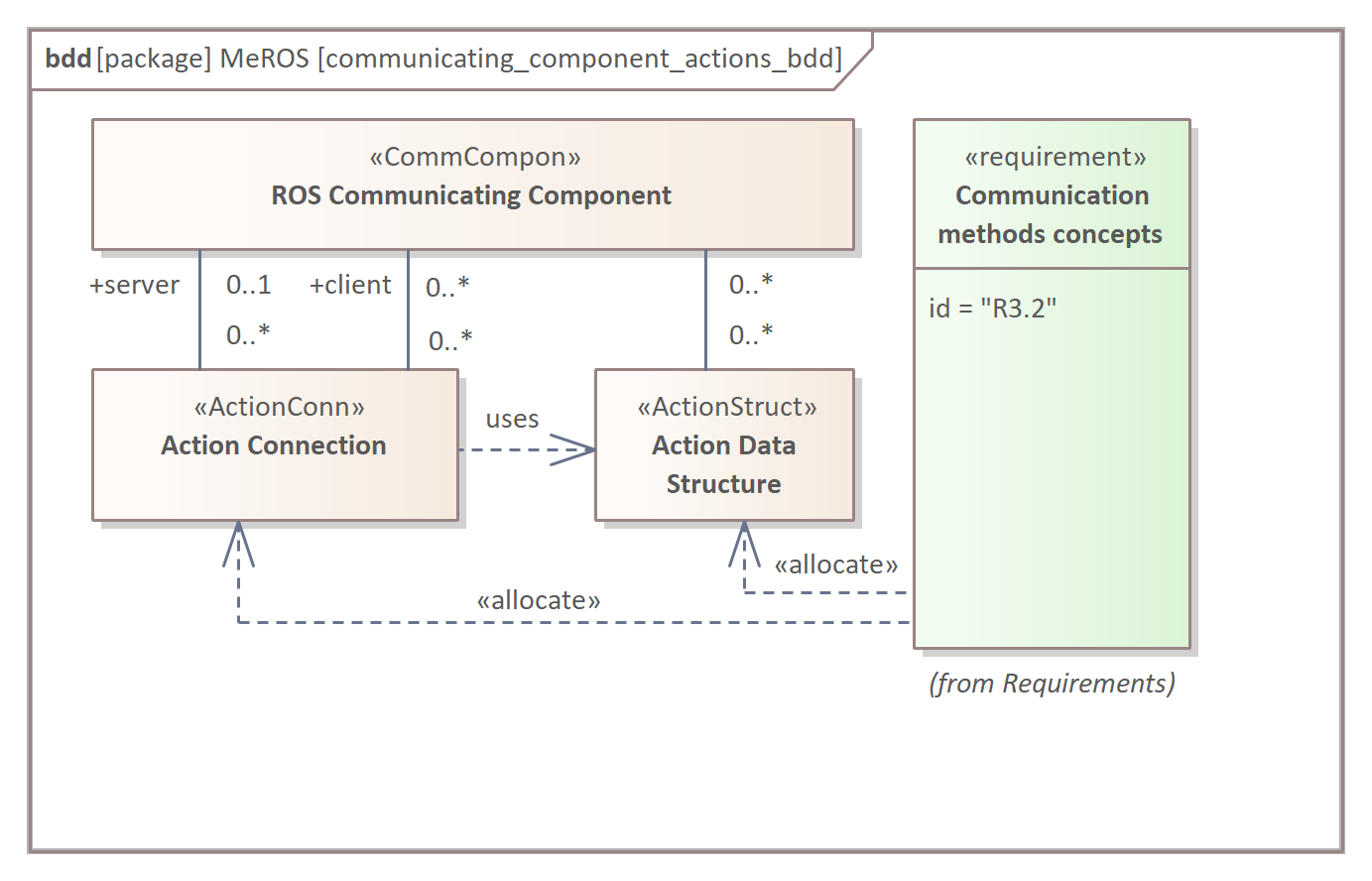}}
		\end{center}
		\caption{Communicating Component relations -- actions -- bdd.} 
		\label{fig:communicating_component_actions_bdd}
	\end{figure}

\begin{figure}[hbt]
		\centering
		\begin{center}
			{\includegraphics[scale=0.68]{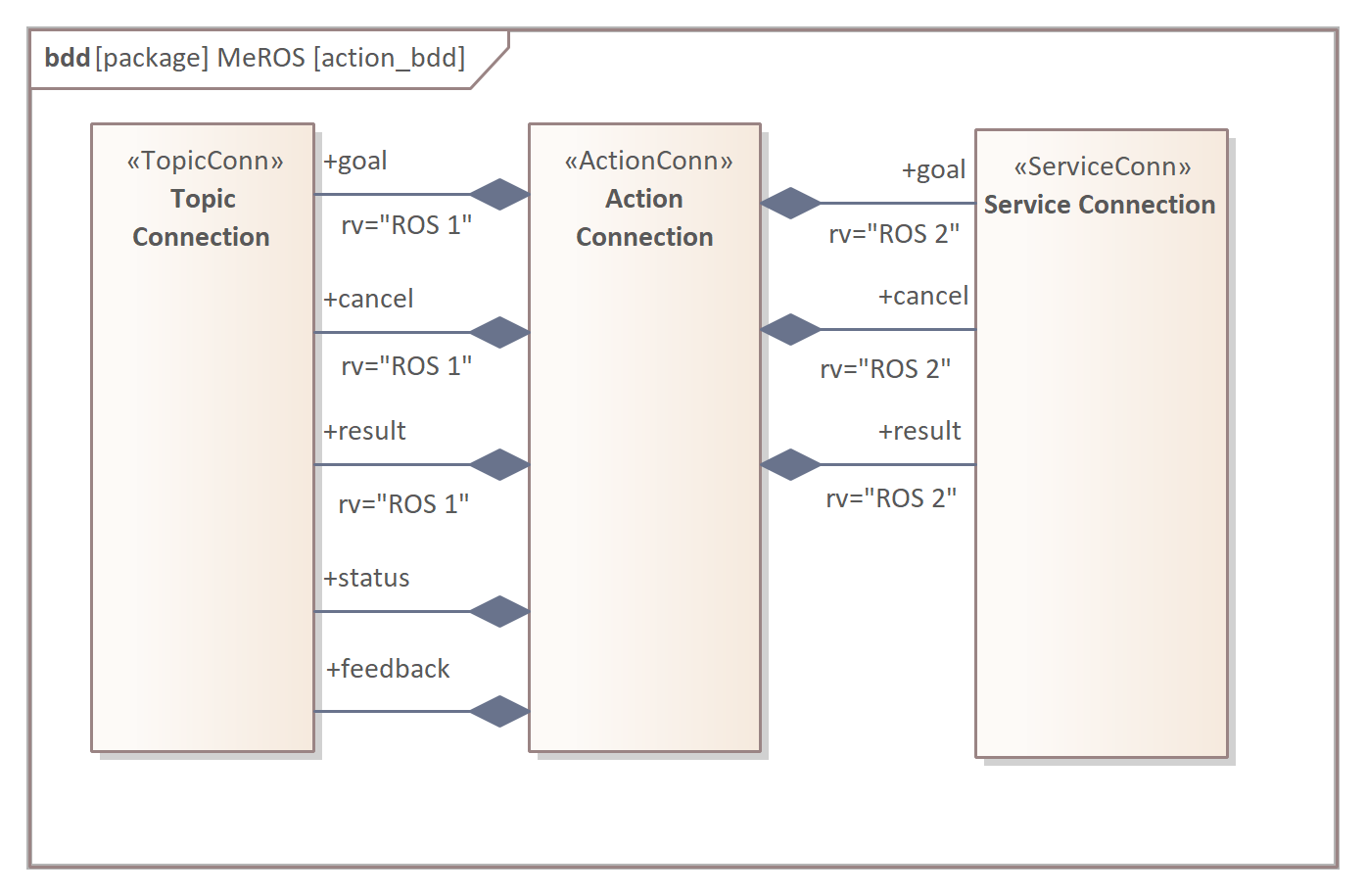}}
		\end{center}
		\caption{Action -- bdd.} 
		\label{fig:action_bdd}
	\end{figure}

	In ROS~2, an Action bases on Topics and Services while in ROS~1 only on Topics. From functional point of view, ROS~1 specific Topics included in Action have their equivalents in ROS~2 specific Services. The Actions are depicted in two diagrams -- Fig.~\ref{fig:communicating_component_actions_bdd} and Fig.~\ref{fig:action_bdd}. Similarly to Services, the Communicating Component can act as a server or a client.	 
	
	% DLA SKLADU ZMIENIONO WIELKOSC Z 0.7
	
\begin{figure}[htb]
		\centering
		\begin{center}
			{\includegraphics[scale=0.73]{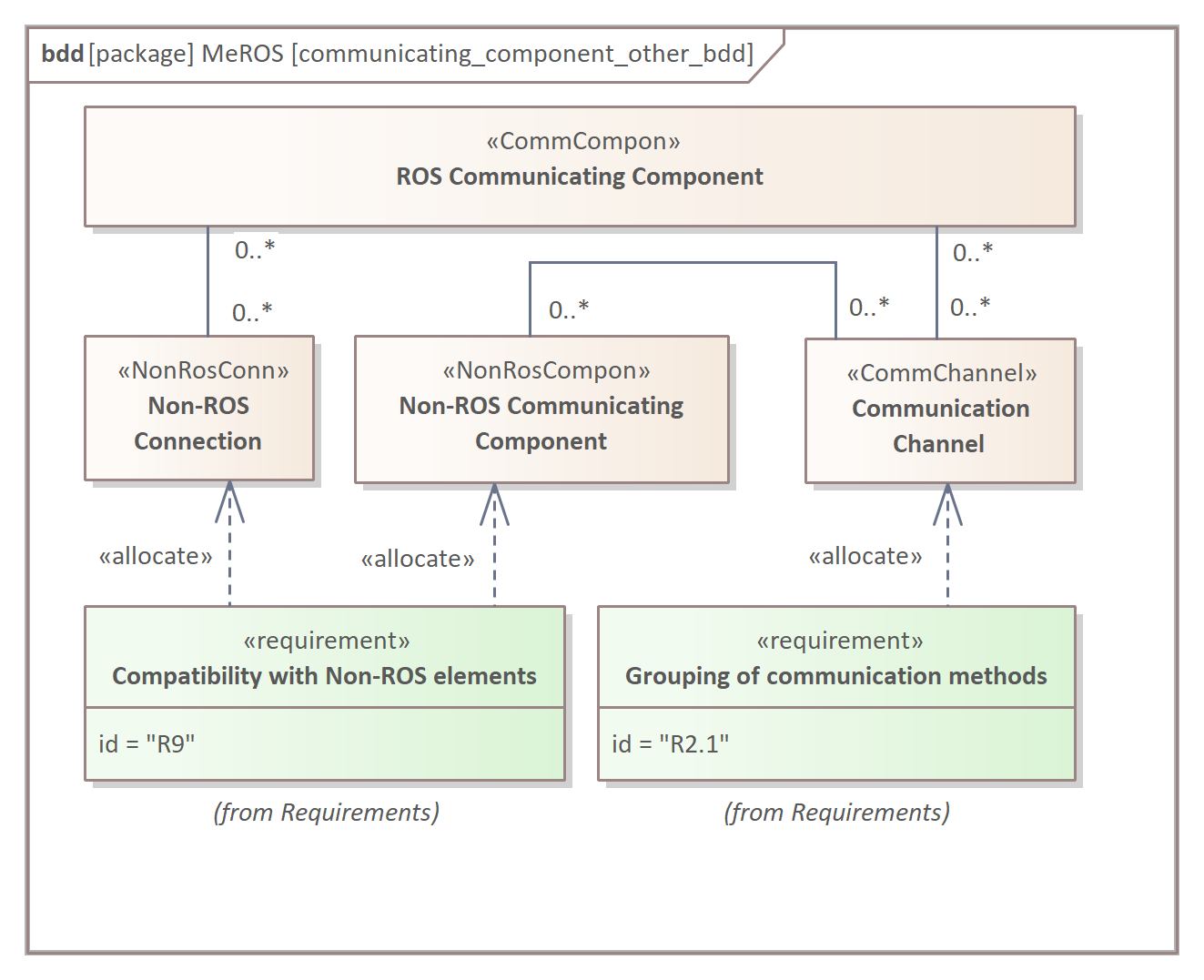}}
		\end{center}
		\caption{Communicating Component relations -- aggregates, Non-ROS elements -- bdd.} 
		\label{fig:communicating_component_other_bdd}
	\end{figure}

	Fig.~\ref{fig:communicating_component_other_bdd} describes how Non-ROS elements are taken into account in relation to communication. Additionally, the figure presents Communication Channel relation to ROS Communicating Component.
	
	% DLA SKLADU ZMIENIONO WIELKOSC Z 0.7	
	
	Besides standard ROS communication methods, the Non-ROS are also included (e.g., http request) to achieve interfaces with Non-ROS parts of the general system. An Action Data Structure comprises data used by three of five Topics composed in Action, i.e., goal, feedback and result. Two remaining Topics, i.e., cancel and status are standardised.
	
\begin{figure}[htb]
	 	\centering
	 	\begin{center}
	 		{\includegraphics[scale=0.73]{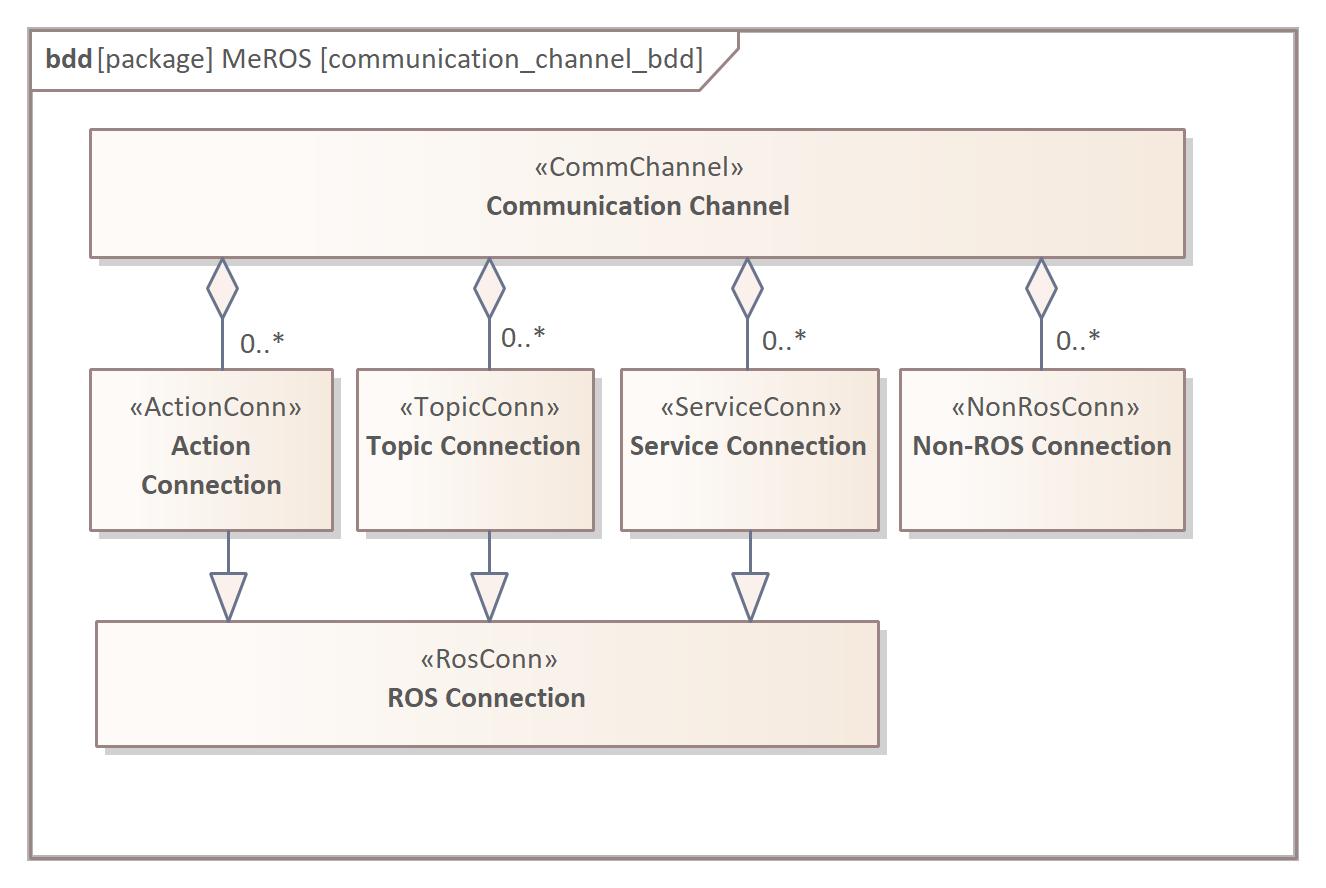}}
	 	\end{center}
	 	\caption{Communication Channel -- bdd.} 
	 	\label{fig:communication_channel_bdd}
	 \end{figure}

	 The Communication Channel \cite{palka2022communication} concept depicted in Fig.~\ref{fig:communication_channel_bdd} is introduced to aggregate specializations of ROS connection (Topic connections, Service connections, and Action connections) as well as Non-ROS Connections. 
	 
  	% DLA SKLADU ZMIENIONO WIELKOSC Z 0.7	
	 
\begin{figure}[htb]
	 	\centering
	 	\begin{center}
	 		{\includegraphics[scale=0.71]{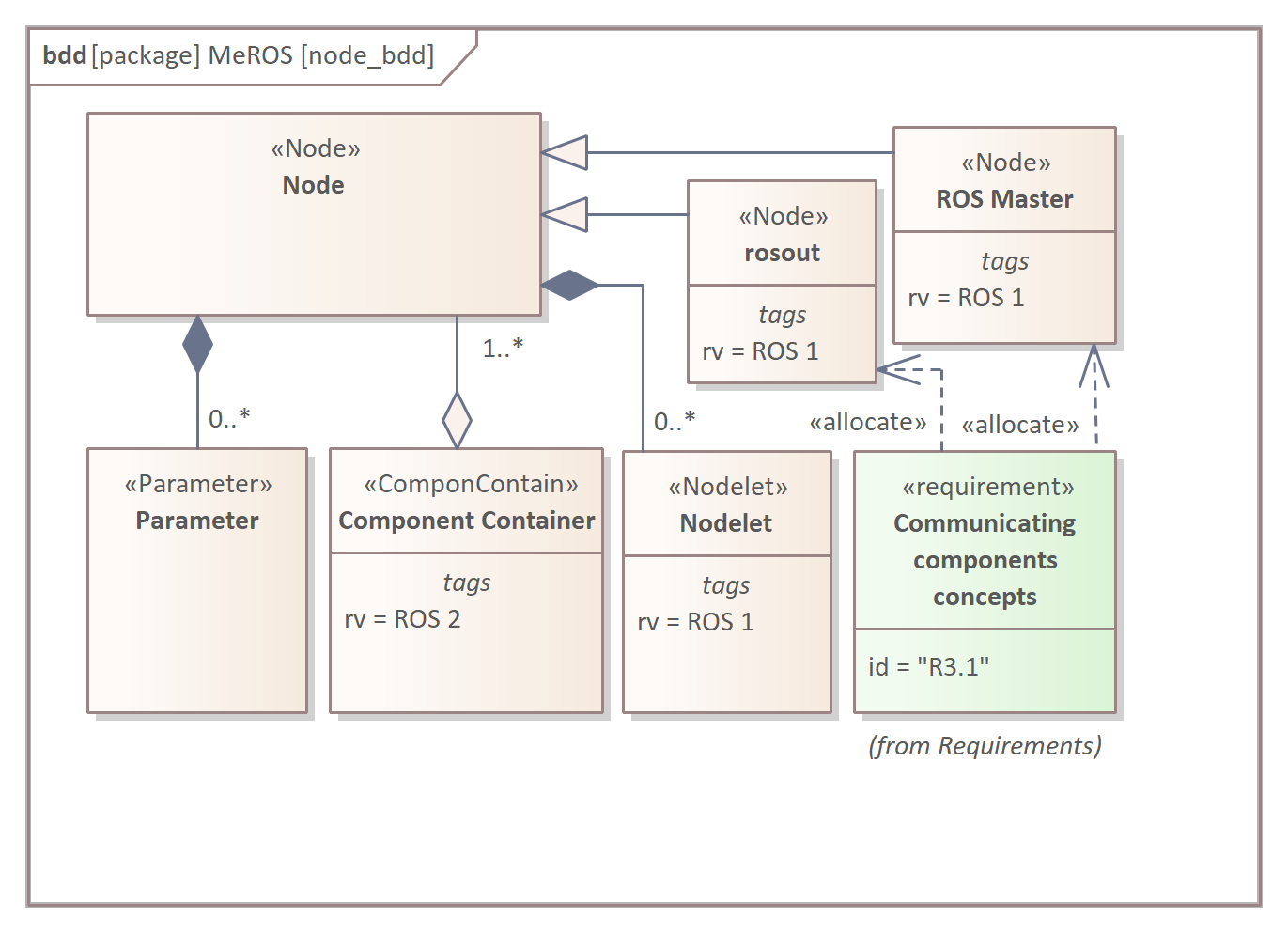}}
	 	\end{center}
	 	\caption{Node -- bdd.} 
		 	\label{fig:node_bdd}
	 \end{figure}

 	The Node (Fig.~\ref{fig:node_bdd}) composes Parameters and Nodelets (the latter in ROS~1). Two specific Nodes are considered in the metamodel: ROS Master and rosout. In ROS~2, Component Container aggregates Nodes executed in a single process.
 	
	 	% DLA SKLADU ZMIENIONO WIELKOSC Z 0.7	
	 
\begin{figure}[htb]
		\centering
		\begin{center}
			{\includegraphics[scale=0.74]{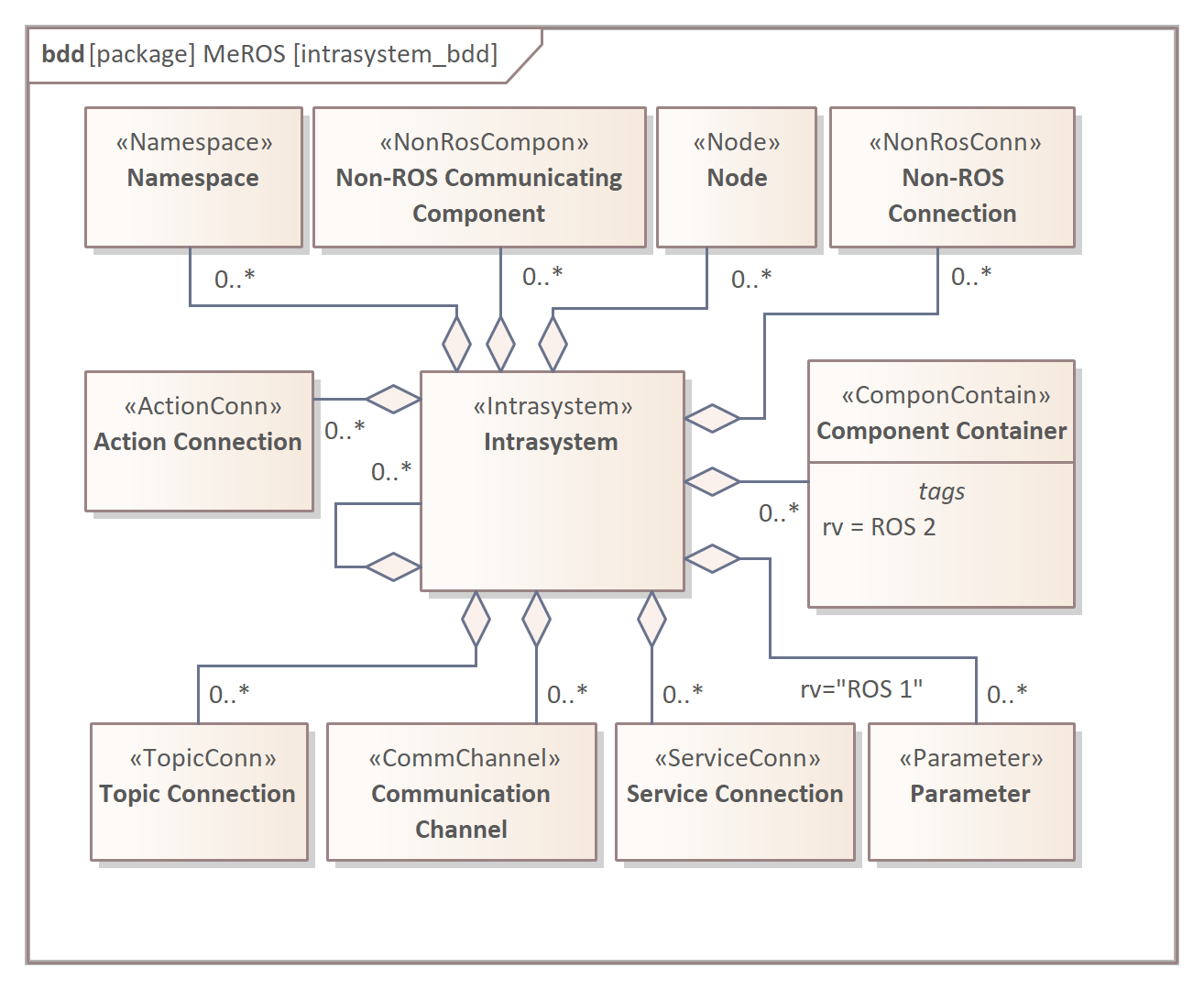}}
		\end{center}
		\caption{Intrasystem compositions -- bdd.} 
		\label{fig:intrasystem_bdd}
	\end{figure}

	The Intrasystem (Fig.~\ref{fig:intrasystem_bdd}) composes ROS and Non-ROS Communicating Component specializations as well as Connections between them. A Parameter block is introduced also for ROS~1. In ROS~2, due to safety reasons, Parameter is composed only into Nodes. Optionally Intrasystem composes the other Intrasystems.
	
	% DLA SKLADU ZMIENIONO WIELKOSC Z 0.7	
	 
\begin{figure}[htb]
		\centering
		\begin{center}
			{\includegraphics[scale=0.73]{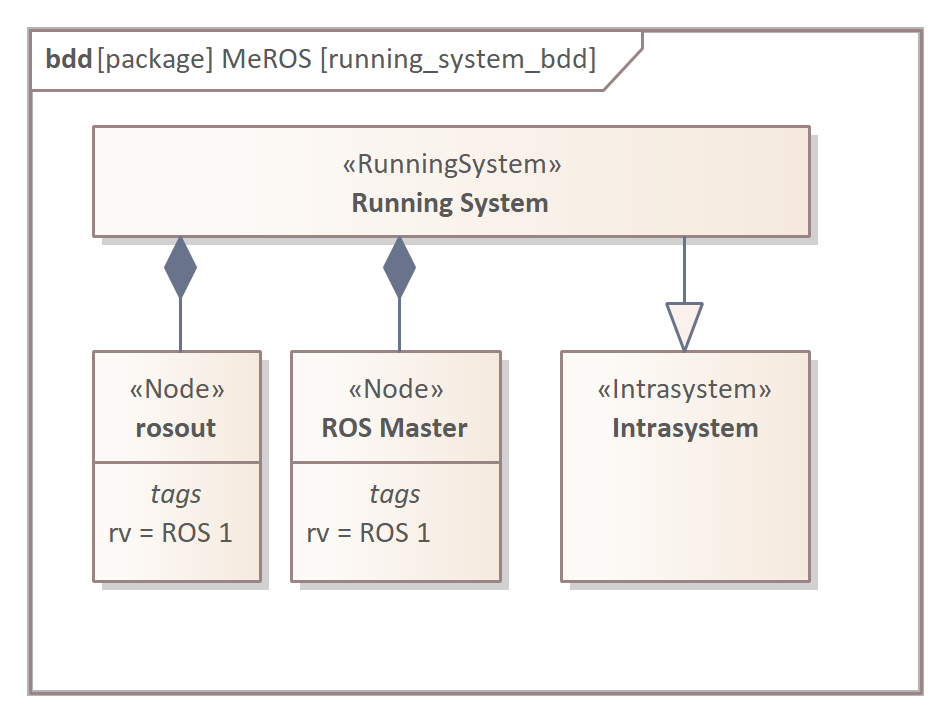}}
		\end{center}
		\caption{Running System compositions -- bdd.} 
		\label{fig:running_system_bdd}
	\end{figure}

  	The Running System (Fig.~\ref{fig:running_system_bdd}) is a specialisation of the Intrasystem that can be executed. Hence, two Nodes are needed for ROS~1: rosout and ROS master. 
	It should be noted that although MeROS could be classified as PSM, the initial, general system description with Communications Channels and Intrasystems corresponds to PIM specification. Then, the detailing of these aggregates corresponds to the transition from PIM to PSM. 
	
			% DLA SKLADU ZMIENIONO WIELKOSC Z 0.7	
	
\begin{figure}[htb]
		\centering
		\begin{center}
			{\includegraphics[scale=0.7]{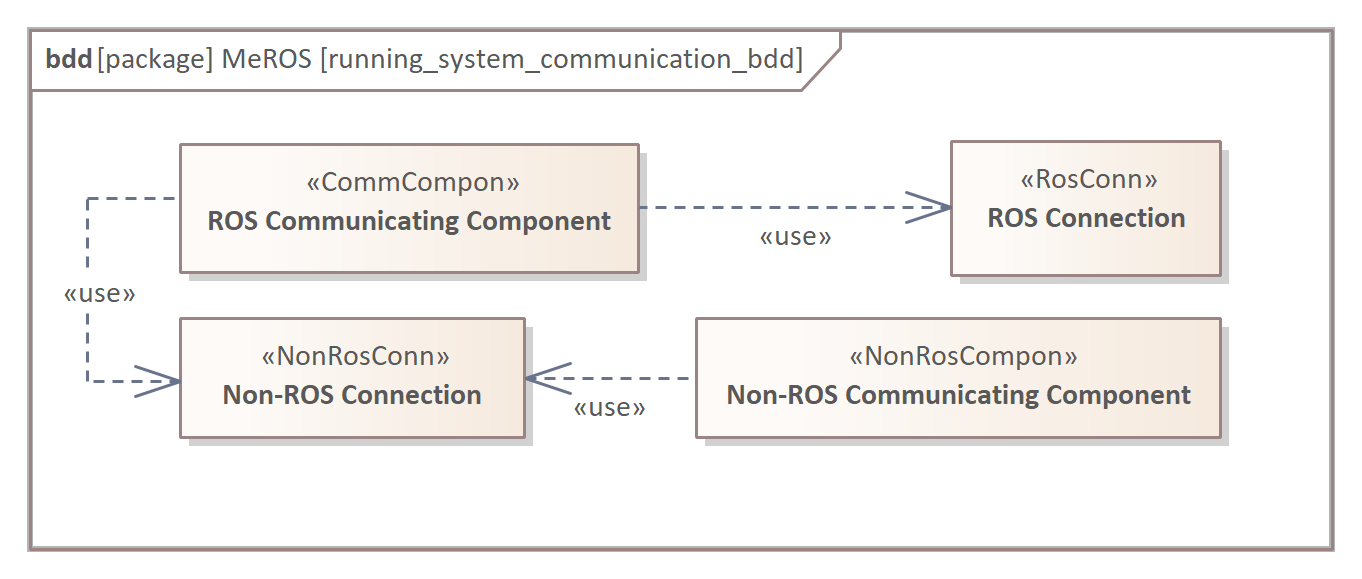}}
		\end{center}
		\caption{Running System communication -- bdd.} 
		\label{fig:running_system_communication_bdd}
	\end{figure}

	The way Communicating Components use various types of connections is presented in Fig.~\ref{fig:running_system_communication_bdd}. Both ROS and Non-ROS Communicating Components can communicate via Non-ROS Connections, but only ROS Communicating Components use ROS Connections.

	% DLA SKLADU ZMIENIONO WIELKOSC Z 0.7	

\begin{figure}[htb]
		\centering
		\begin{center}
			{\includegraphics[scale=0.75]{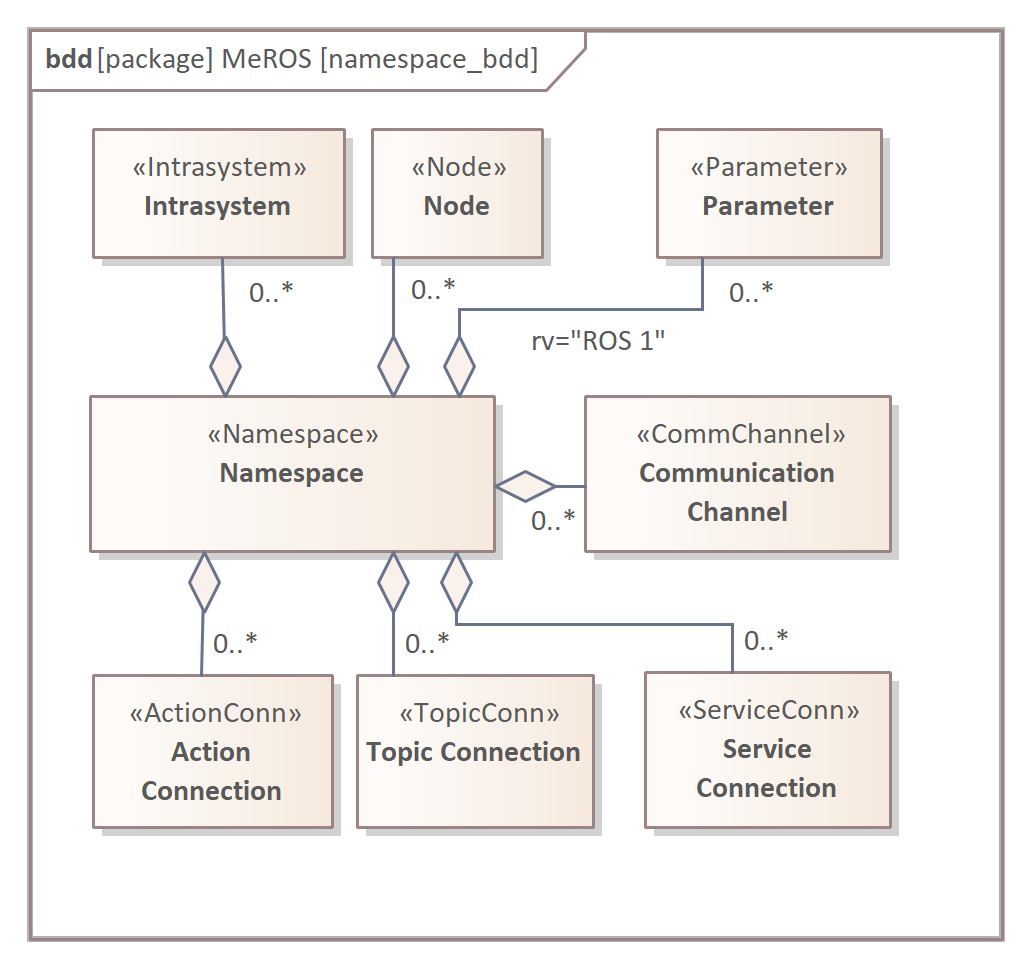}}
		\end{center}
		\caption{Namespace composition -- bdd.} 
		\label{fig:namespace_bdd}
	\end{figure}

	The Namespace (Fig.~\ref{fig:namespace_bdd}) aggregates elements of the Intrasystem, but only ROS related. In opposition to the Intrasystem, the Namespace does not specialise Communicating Component. Hence, it can not act as Communicating Component.
	
		% DLA SKLADU ZMIENIONO WIELKOSC Z 0.7	
	
\begin{figure}[htb]
		\centering
		\begin{center}
			{\includegraphics[scale=0.7]{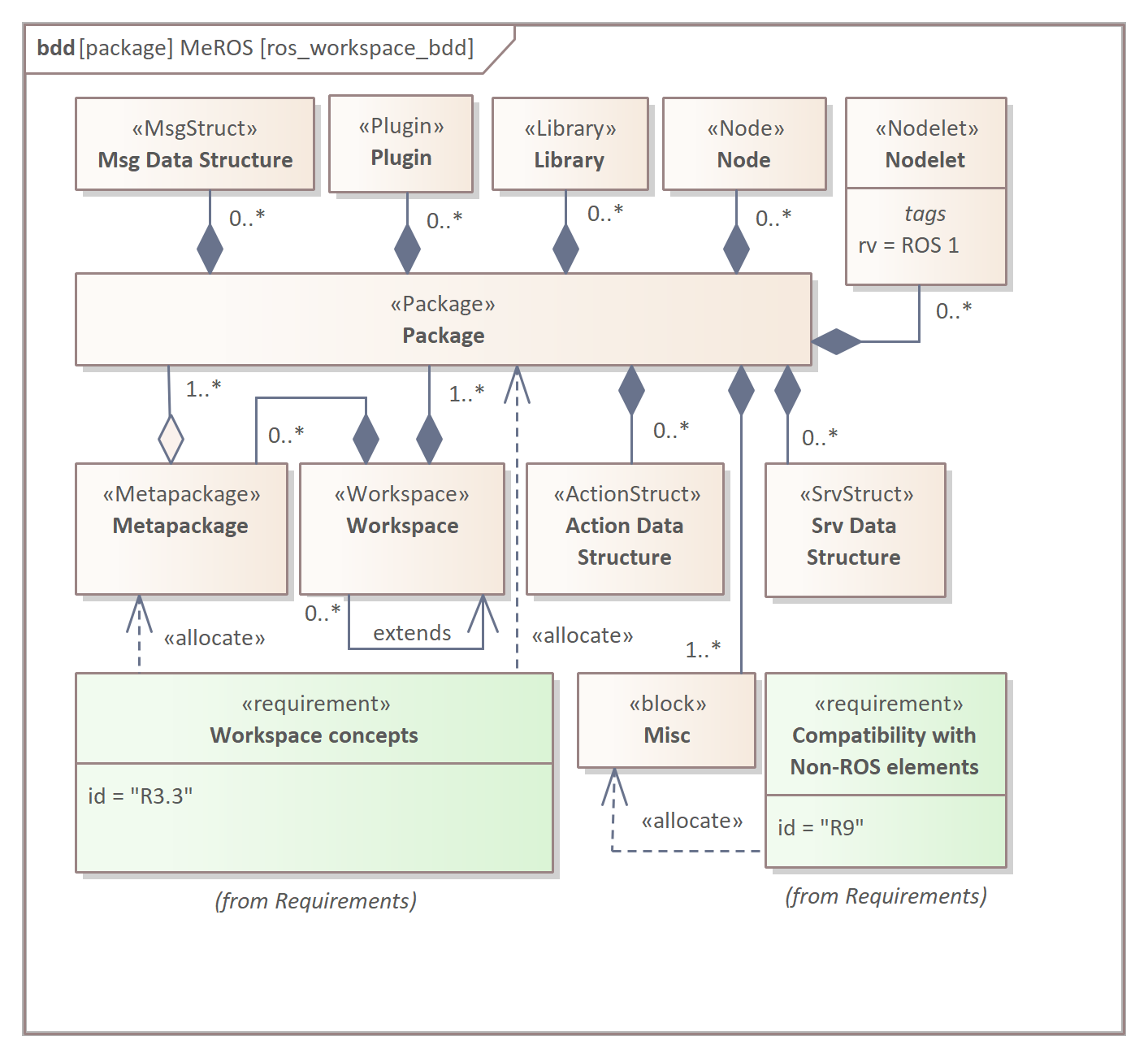}}
		\end{center}
		\caption{ROS Workspace composition -- bdd.} 
		\label{fig:ros_workspace_bdd}
	\end{figure}

	The Workspace (Fig.~\ref{fig:ros_workspace_bdd}) contains of Packages that compose the files related to general ROS concepts such as Node source codes, communication structures definitions, etc. 
	
	% DLA SKLADU ZMIENIONO WIELKOSC Z 0.7	
	
	It should be noted that in case of Actions, specific communication structures definitions are stored in Action Data Structures. The Misc <<block>> relates to other ROS and Non-ROS files, e.g., roslaunch configuration, obligatory package.xml, CMakeLists.txt. The Metapackage is introduced for conformity with the latest ROS~1 releases [R7] as well as ROS~2.

\subsection{Communication}
\label{sec:metamodel-communication}
	
	This section depicts the behavioural and structural aspects of communication in the system. The previous section considers block definition diagrams (bdd). In the following part, the internal block diagrams (ibd) and behavioural diagrams are discussed. The goal is to present three modes of communication: Topic [R4.1] (sec.~\ref{sec:metamodel-topic}), Service [R4.2] (sec.~\ref{sec:metamodel-service}) and Action [R4.3] (sec.~\ref{sec:metamodel-action}). It should be noted that the concept of presentation of communication with and without a dedicated communication component is illustrated on communication with Topics but can also be applied to Services, Actions and Communication Channels.
	
	\subsubsection{Topic}
	\label{sec:metamodel-topic}
		
\begin{figure}[htb]
		\centering
		\begin{center}
			{\includegraphics[scale=0.8]{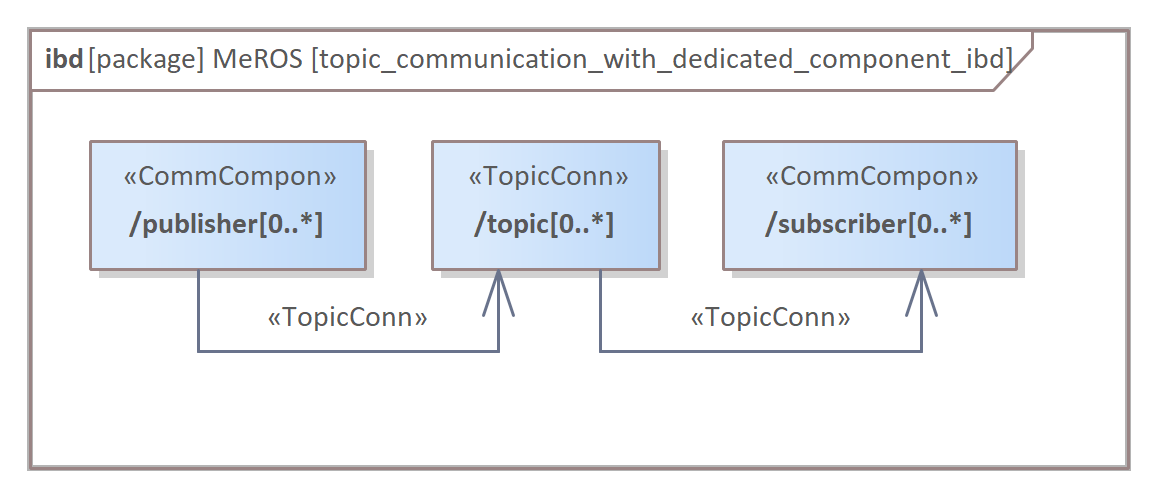}}
		\end{center}
		\caption{Topics with dedicated communication components -- all components -- ibd.} 
		\label{fig:topic_communication_with_dedicated_component_ibd}
	\end{figure}

	Fig.~\ref{fig:topic_communication_with_dedicated_component_ibd} presents the ibd diagram of publishers' and subscribers' communication via topics. This diagram uses a dedicated communication component for each Topic [R4.1.1]. There are no general limits to the number of publishers, subscribers and Topics they communicate with. 
	
	% DLA SKLADU ZMIENIONO WIELKOSC Z 0.7	

\begin{figure}[htb]
		\centering
		\begin{center}
			{\includegraphics[scale=0.9]{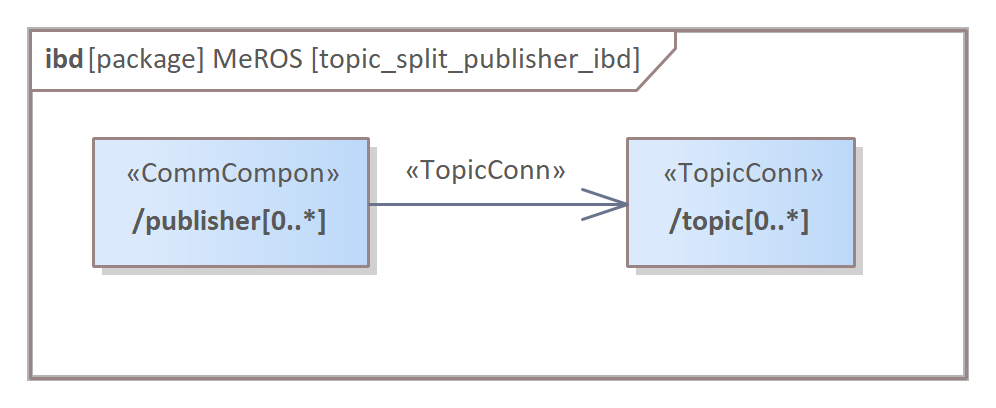}}
		\end{center}
		\caption{Topics with dedicated communication components -- publisher -- ibd.} 
		\label{fig:topic_split_publisher_ibd}
	\end{figure}

\begin{figure}[htb]
		\centering
		\begin{center}
			{\includegraphics[scale=0.9]{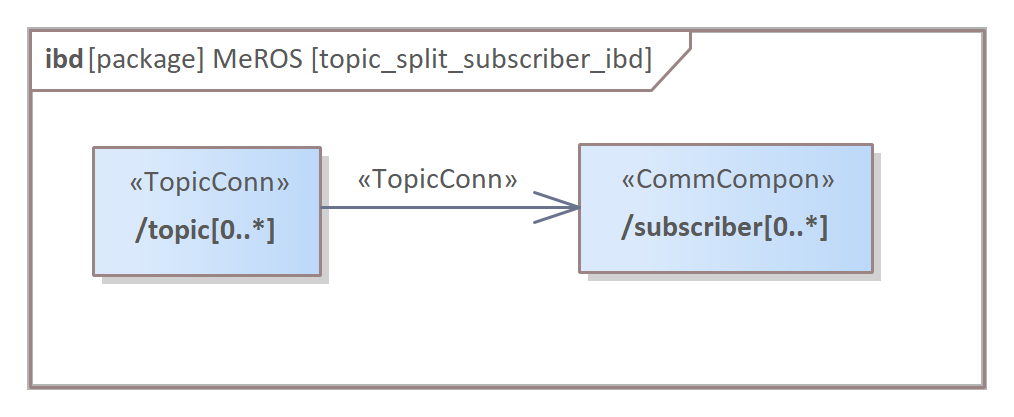}}
		\end{center}
		\caption{Topics with dedicated communication components -- subscriber -- ibd.} 
		\label{fig:topic_split_subscriber_ibd}
	\end{figure}

	Thanks to a dedicated component to represent communication, the diagram in Fig.~\ref{fig:topic_communication_with_dedicated_component_ibd} can be split into two considering publisher (Fig.~\ref{fig:topic_split_publisher_ibd}) and subscriber (Fig.~\ref{fig:topic_split_subscriber_ibd}) separately, without losing information. It is especially useful when system fragments are presented after its decomposition that subdivides communication channels.
	
	% DLA SKLADU ZMIENIONO WIELKOSC Z 0.7	

	% DLA SKLADU ZMIENIONO WIELKOSC Z 0.7	
	
\begin{figure}[htb]
		\centering
		\begin{center}
			{\includegraphics[scale=0.73]{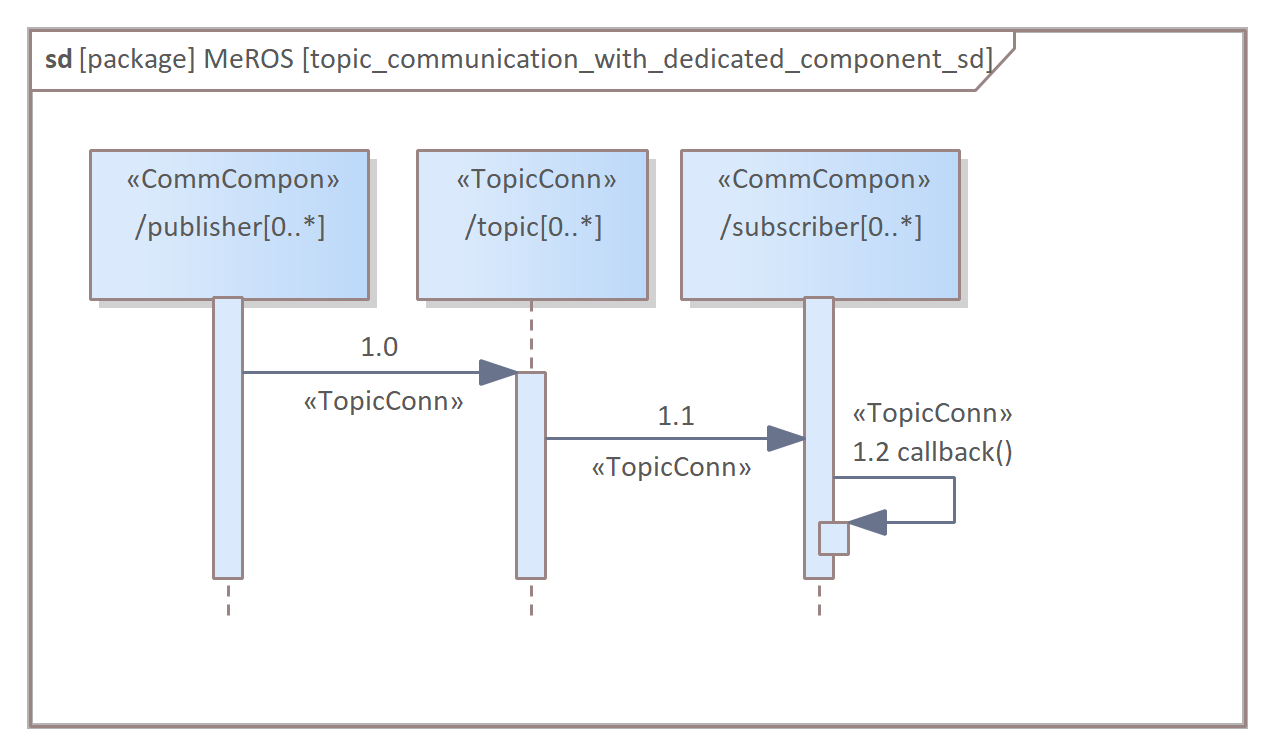}}
		\end{center}
		\caption{Topics with dedicated communication components -- sd.} 
		\label{fig:topic_communication_with_dedicated_component_sd}
	\end{figure}

\begin{figure}[htb]
		\centering
		\begin{center}
			{\includegraphics[scale=0.78]{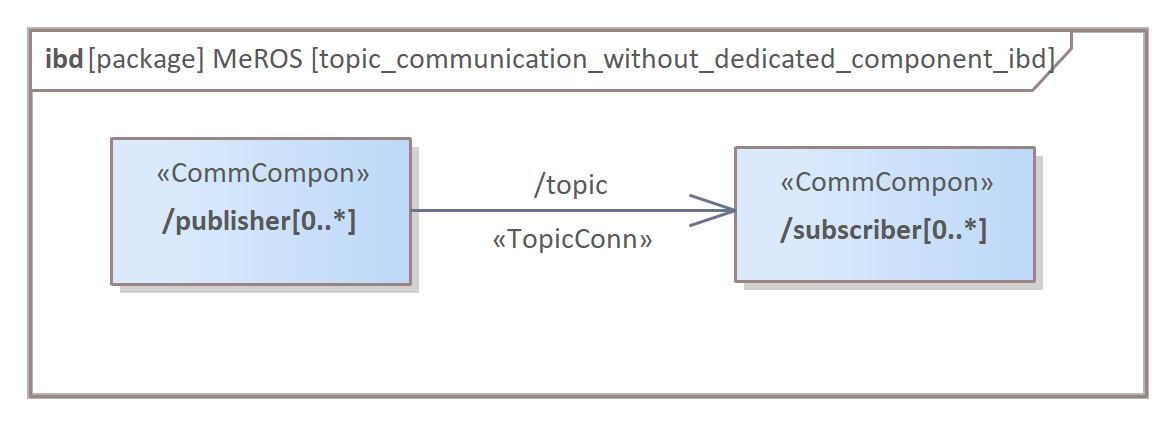}}
		\end{center}
		\caption{Topics without dedicated communication components -- ibd.} 
		\label{fig:topic_communication_without_dedicated_component_ibd}
	\end{figure}

\begin{figure}[H]
		\centering
		\begin{center}
			{\includegraphics[scale=0.78]{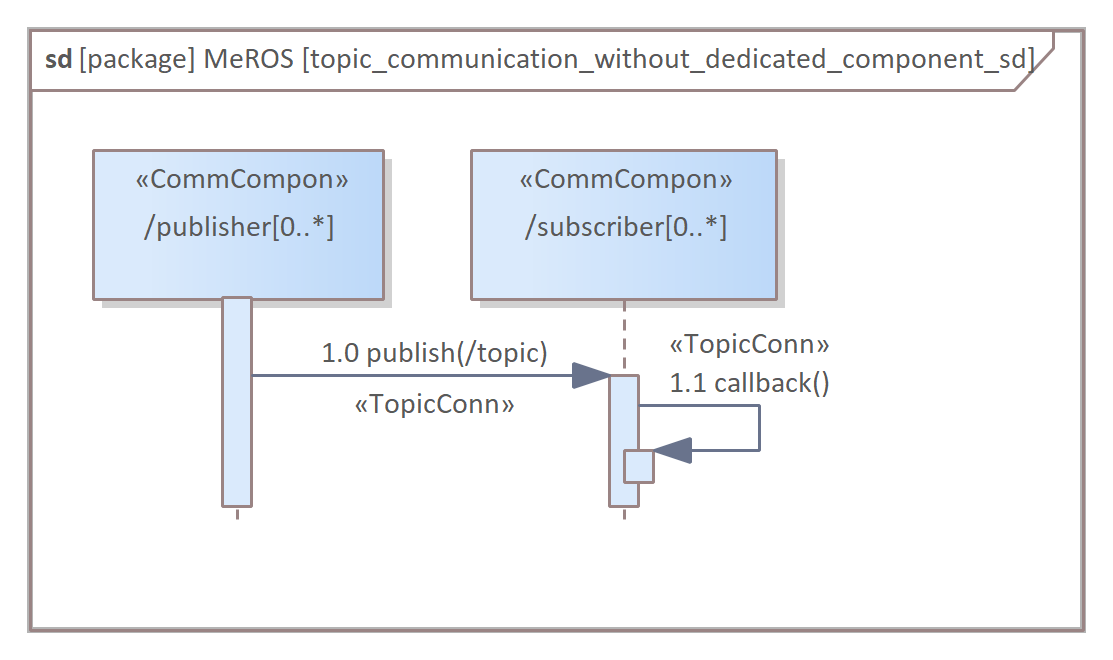}}
		\end{center}
		\caption{Topics without dedicated communication components -- sd.} 
		\label{fig:topic_communication_without_dedicated_component_sd}
	\end{figure}

	Fig. \ref{fig:topic_communication_with_dedicated_component_sd} depicts the corresponding sequence diagram. Publishers send a message through Topics to the subscribers. The incoming message cause the subscriber to execute the callback function. 	Fig.~\ref{fig:topic_communication_without_dedicated_component_ibd} and Fig.~\ref{fig:topic_communication_without_dedicated_component_sd} present an alternative approach to depict the system communicating via topics. In this case, no dedicated communication components are used [R4.1.2]. 
	
\subsubsection{Service}
\label{sec:metamodel-service}
	
\begin{figure}[htb]
		\centering
		\begin{center}
			{\includegraphics[scale=0.73]{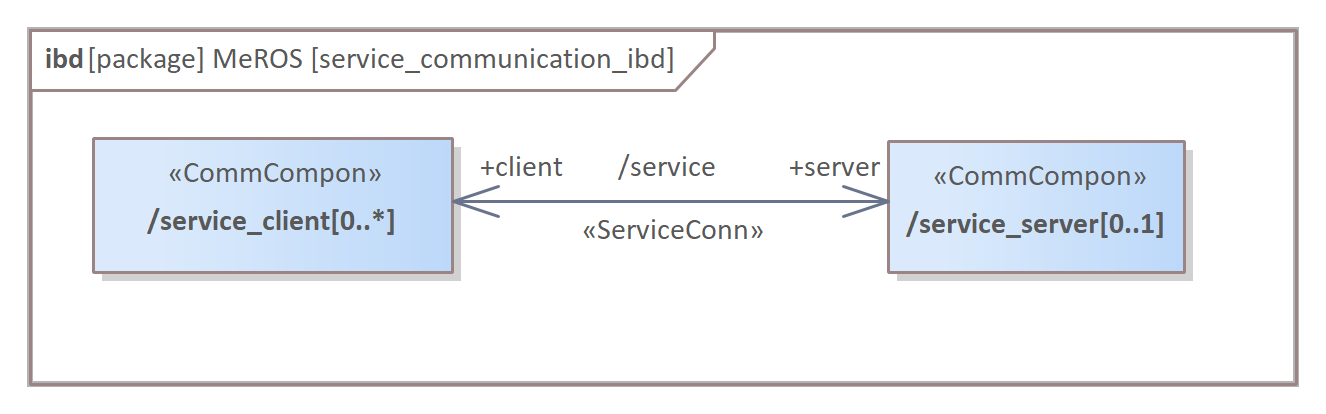}}
		\end{center}
		\caption{Service-based communication -- ibd.} 
		\label{fig:service_communication_ibd}
	\end{figure}

\begin{figure}[htb]
		\centering
		\begin{center}
			{\includegraphics[scale=0.85]{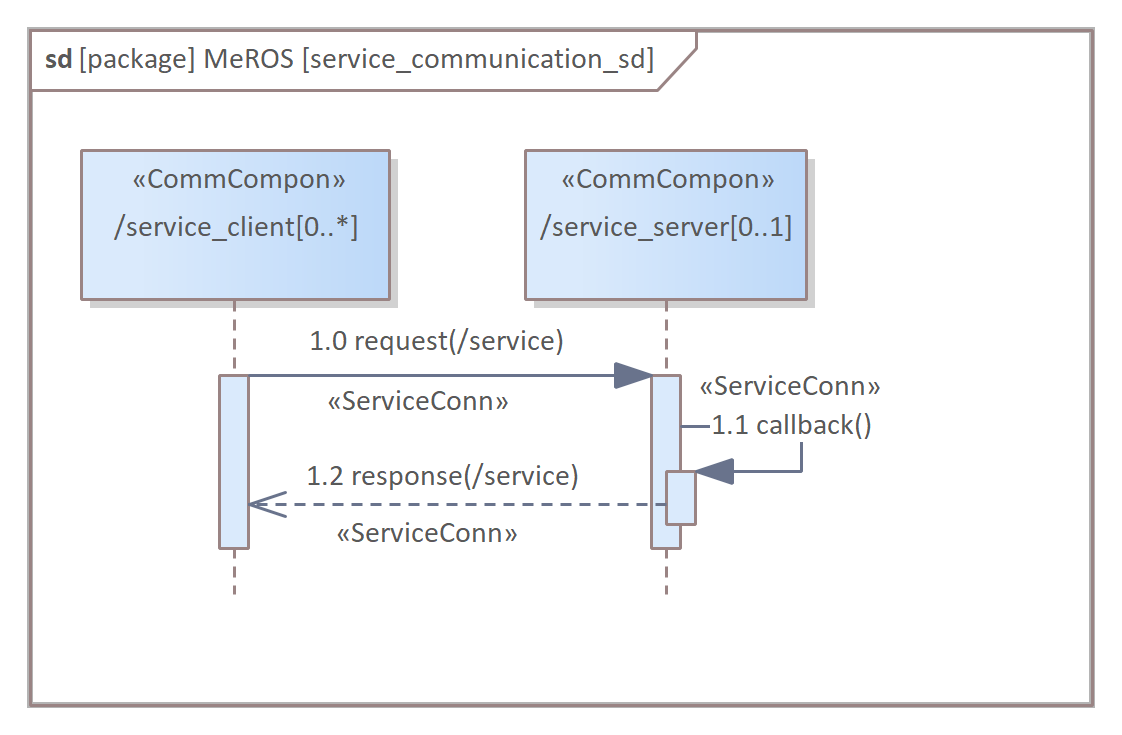}}
		\end{center}
		\caption{Service-based communication -- sd.} 
		\label{fig:service_communication_sd}
	\end{figure}

	For each ROS Service, there is at most one server and a~number of clients (Fig.~\ref{fig:service_communication_ibd} and Fig.~\ref{fig:service_communication_sd}). Service-type communication is bidirectional and realises RPC (remote procedure call).
	
	% DLA SKLADU ZMIENIONO WIELKOSC Z 0.7	
	
	% DLA SKLADU ZMIENIONO WIELKOSC Z 0.7	

\subsubsection{Action}
\label{sec:metamodel-action}
	
\begin{figure}[htb]
		\centering
		\begin{center}
			{\includegraphics[scale=0.75]{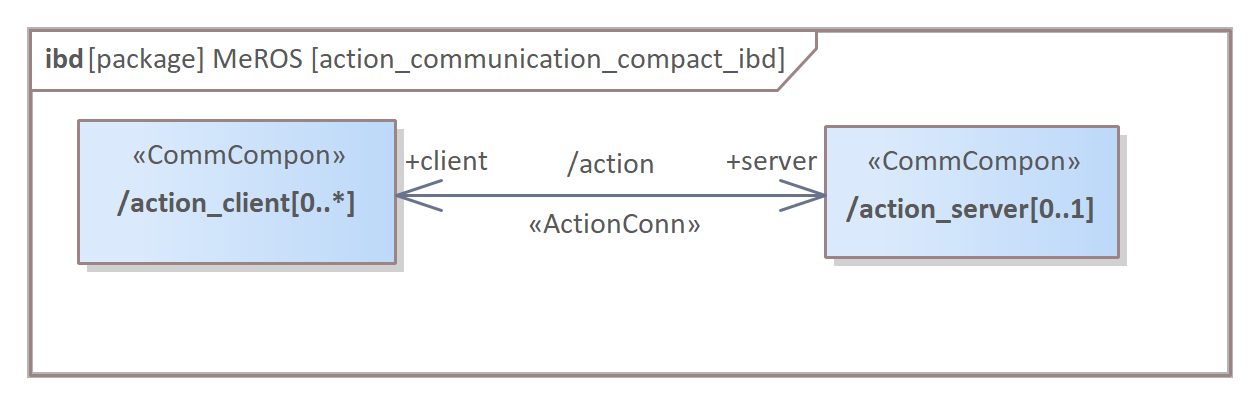}}
		\end{center}
		\caption{Action-based communication -- compact representation -- ibd.} 
		% https://docs.ros.org/en/foxy/Tutorials/Beginner-CLI-Tools/Understanding-ROS2-Actions/Understanding-ROS2-Actions.html
		\label{fig:action_communication_compact_ibd}
	\end{figure}

	ROS Action communication's general, simplified structure (Fig.~\ref{fig:action_communication_compact_ibd}) is analogous to ROS Service. These type of presentation is universal for ROS~1 and ROS~2.
	
	% DLA SKLADU ZMIENIONO WIELKOSC Z 0.7	
	
\begin{figure}[htb]
		\centering
		\begin{center}
			{\includegraphics[scale=0.7]{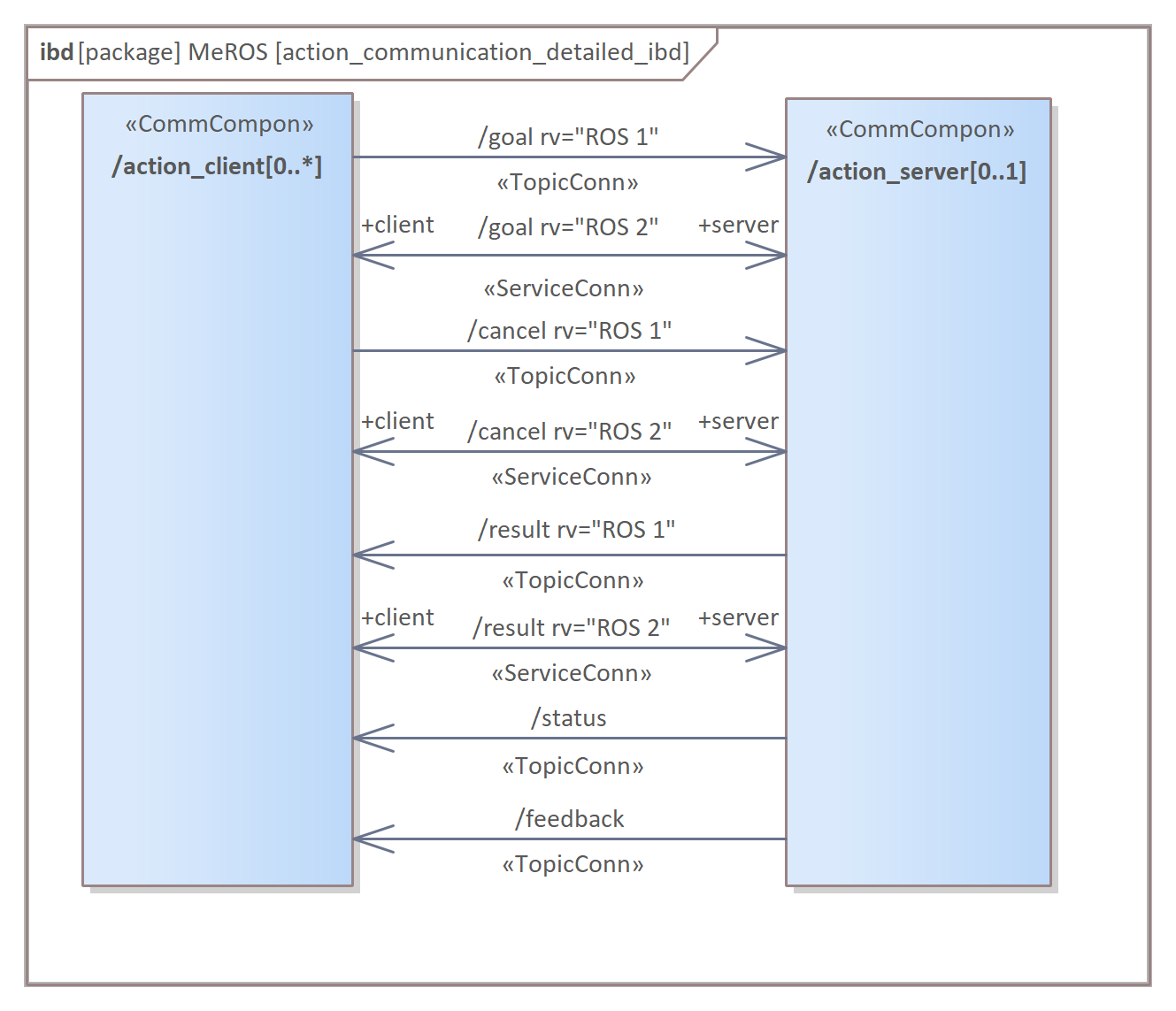}}
		\end{center}
		\caption{Action-based communication -- detailed -- ibd.} 
		% https://design.ros2.org/articles/actions.html
		\label{fig:action_communication_detailed_ibd}
	\end{figure}

	 An Action (Fig.~\ref{fig:action_communication_detailed_ibd}) is based on several Topics in ROS~1, while on Topics and Services in ROS~2.
	
	% DLA SKLADU ZMIENIONO WIELKOSC Z 0.7	
	
\begin{figure}[htb]
		\centering
		\begin{center}
			{\includegraphics[scale=0.84]{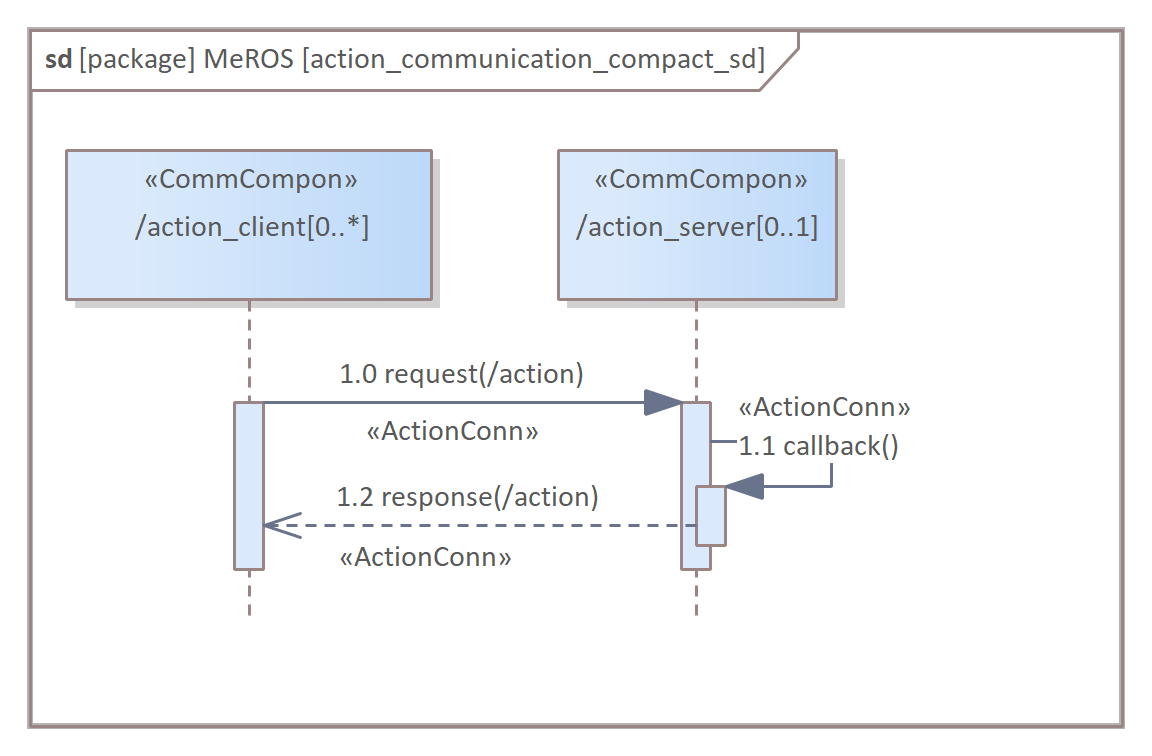}}
		\end{center}
		\caption{Action-based communication sequence -- compact presentation -- sd.} 
		\label{fig:action_communication_compact_sd}
	\end{figure}

	In practice, to present an action-related communication compactly on sd diagram (Fig.~\ref{fig:action_communication_compact_sd}) particular Topics and Services can be generalised as a~request (for /goal and /cancel) and a~response (for /status, /feedback and /result). It should be noted that this diagram presents the Action communication sequence in a simplified way.
	
\begin{figure}[htb]
		\centering
		\begin{center}
			{\includegraphics[scale=0.7]{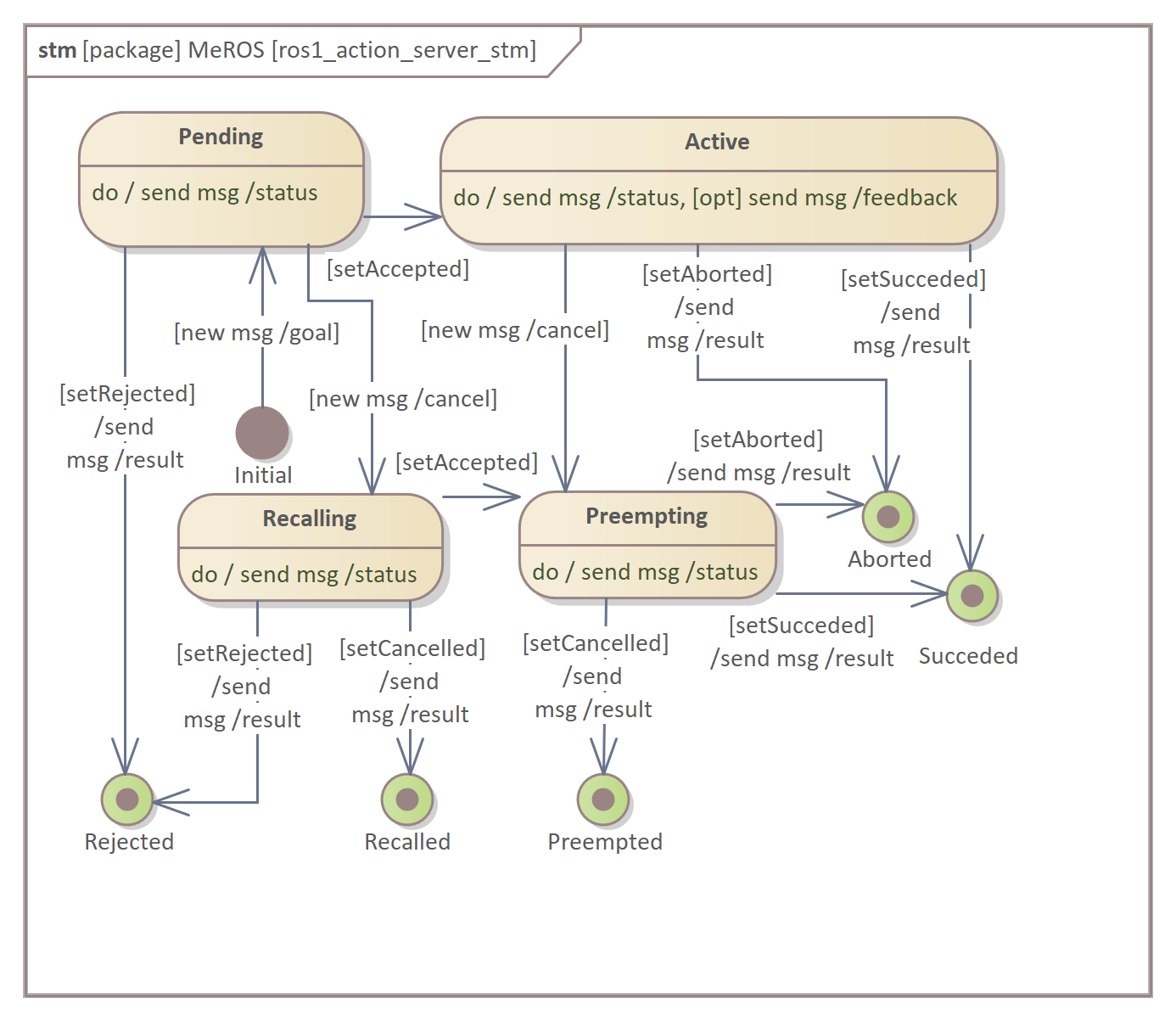}}
		\end{center}
		\caption{ROS~1 Action server -- stm.} 
		\label{fig:ros1_action_server_stm}
	\end{figure}

	The detailed behaviour of the Action server and Action client in ROS~1 is specified by state machines \footnote{\url{http://wiki.ros.org/actionlib/DetailedDescription}}. ROS~2 Action server and Action client behaviour is analogous. Here, these two state machines are depicted in stm diagrams. In the description, in addition to the original ROS wiki presentation, the Topics are directly mentioned both in transitions and states actions.
	 Fig.~\ref{fig:ros1_action_server_stm} depicts the ROS~1 Action server state machine.
	  Its transitions depend on the new messages sent by the Action client or internal predicates. 
	  
\begin{figure}[htb]
		\centering
		\begin{center}
			{\includegraphics[scale=0.7]{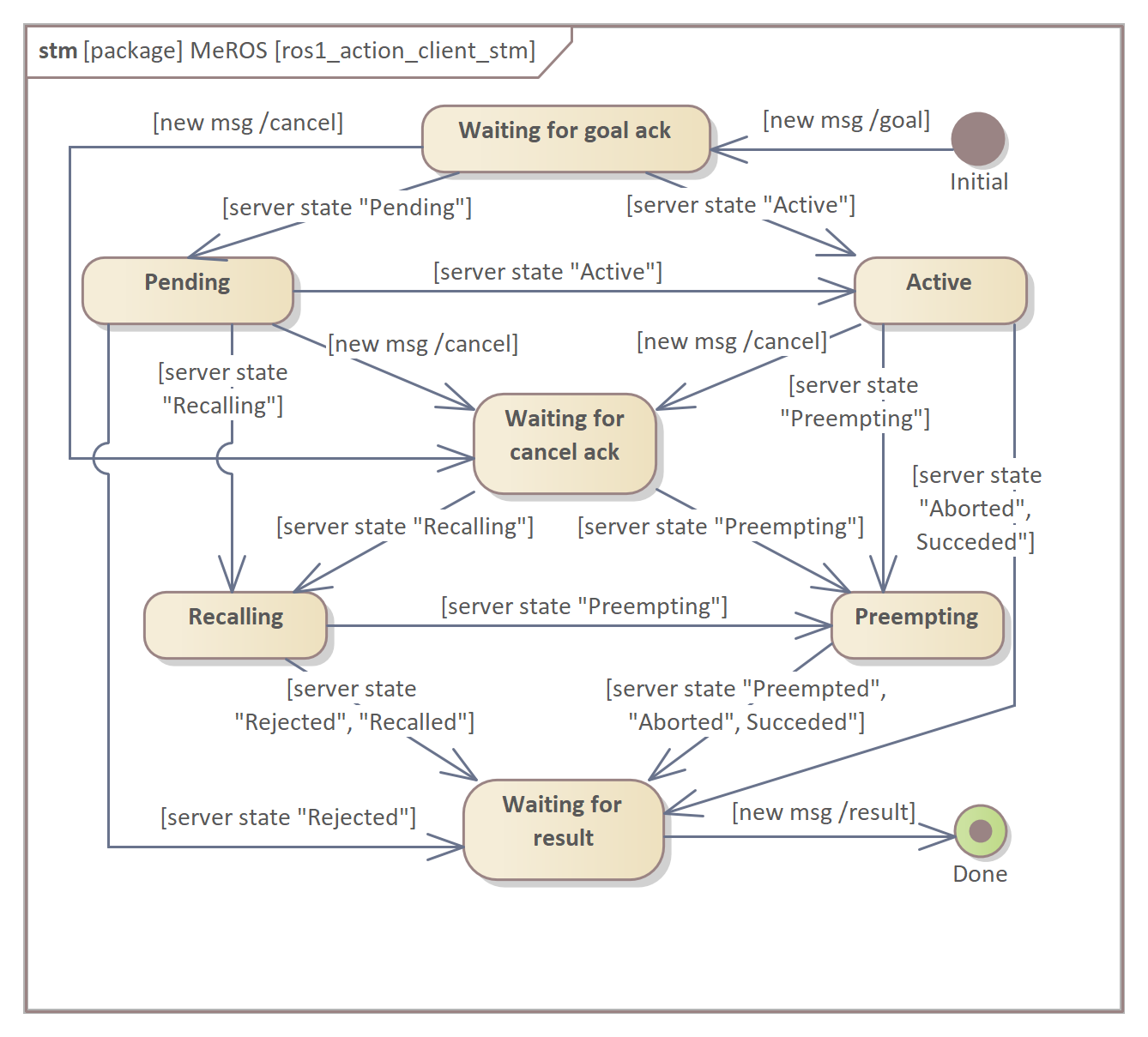}}
		\end{center}
		\caption{ROS~1 Action client -- stm.}
		\label{fig:ros1_action_client_stm}
	\end{figure}

	The ROS~1 Action client state machine (Fig.~\ref{fig:ros1_action_client_stm}) depends on the server state provided by the Action server in /status Topic and internal predicates. 
		
\section{MeROS application}
\label{sec:application}

\subsection{Application hints}
\label{sec:application-hints}

	MeROS metamodel can be employed in various ways in broad context of SE. Although, it is difficult to speak of an indication of the best procedure for its application, it is possible to formulate some practical guidelines for building a particular system model based on MeROS. 

	\begin{itemize}
		\item Before defining a SysML Object, one must define the Block of which it is an instance. It is best to place Block definitions on the bdd diagrams as well. Afterwards, the definition of Objects and the formulation of the other diagrams can follow.
		\item The Object is an instance of the Block, and the Object's classifier corresponds to the Block's name. The Object's name specifies the name of the Block instance. The stereotypes for Block and Object are the same.
		\item The same Blocks and the same Objects should not be duplicated. A Block or Object is defined once and used in different diagrams (in particular, the same Blocks in both the Running System and Workspace diagrams or Objects in the ibd and sd diagrams).
		\item In practice, as long as automatic validation of models formulated in MeROS is not planned, there is no need to formulate a complete model in a SysML project. 
	\end{itemize}
		
	To help develop user projects, the MeROS UML profile and other materials are accessible from MeROS project page\footnote{\url{http://github.com/twiniars/meros}}.

\subsection{Exemplary system}
\label{sec:application-example}

\begin{figure}[htb]
		\centering
		\begin{center}
			{\includegraphics[width=.9\columnwidth]{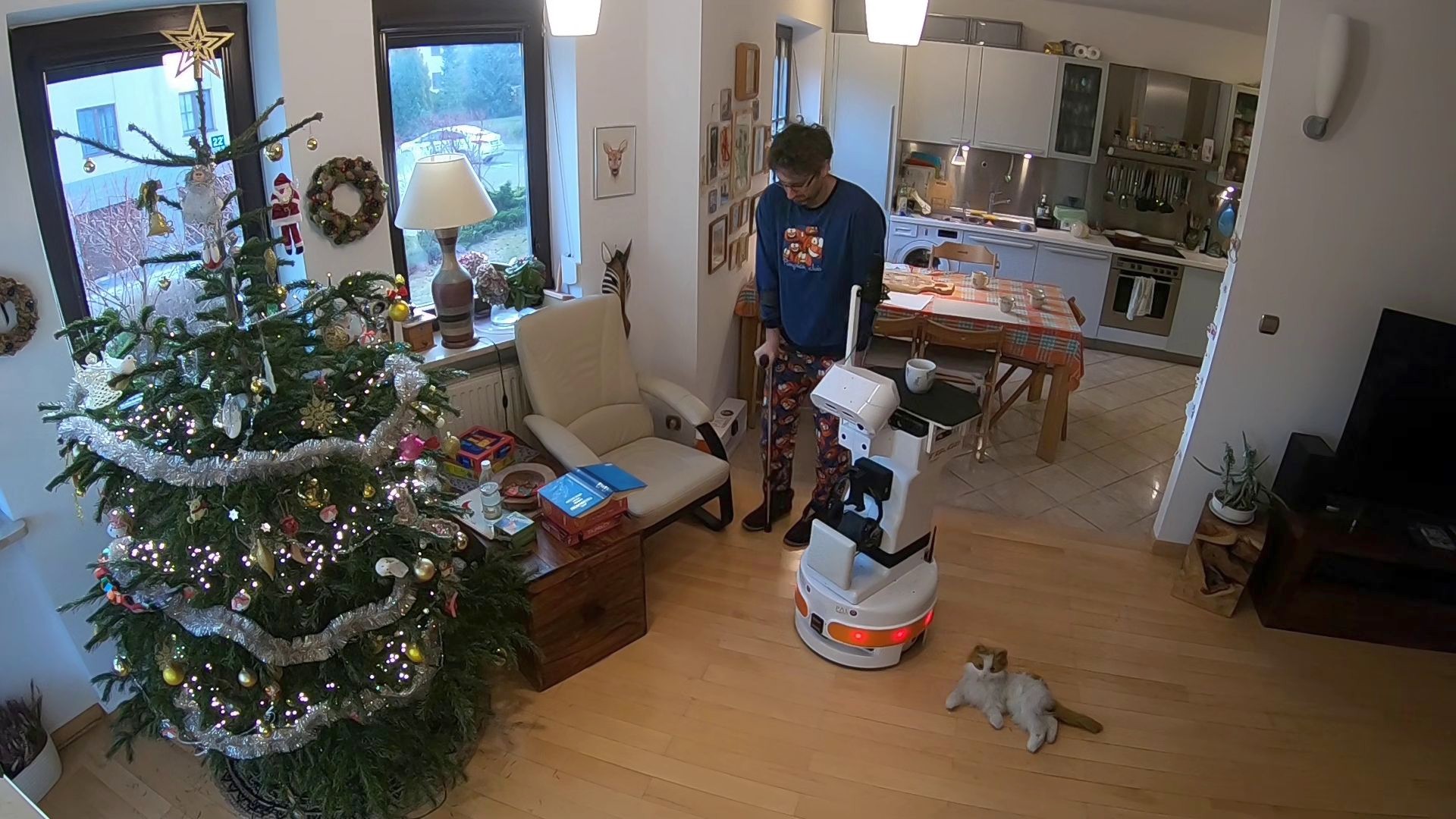}}
		\end{center}
		\caption{Transportation attendance by Rico robot \url{https://vimeo.com/670252925}} 
		\label{fig:herbatka_u_winiara}
	\end{figure}

	This section presents key aspects of an exemplary system development process incorporating MeROS. The exemplary system was created within the AAL INCARE project to control the Rico assistive robot (modified TIAGo platform with controller based on ROS~1) to execute transportation attendance tasks (Fig.~\ref{fig:herbatka_u_winiara}).
	
\begin{figure}[htb] 
		\centering
		\begin{center}
			{\includegraphics[scale=0.67]{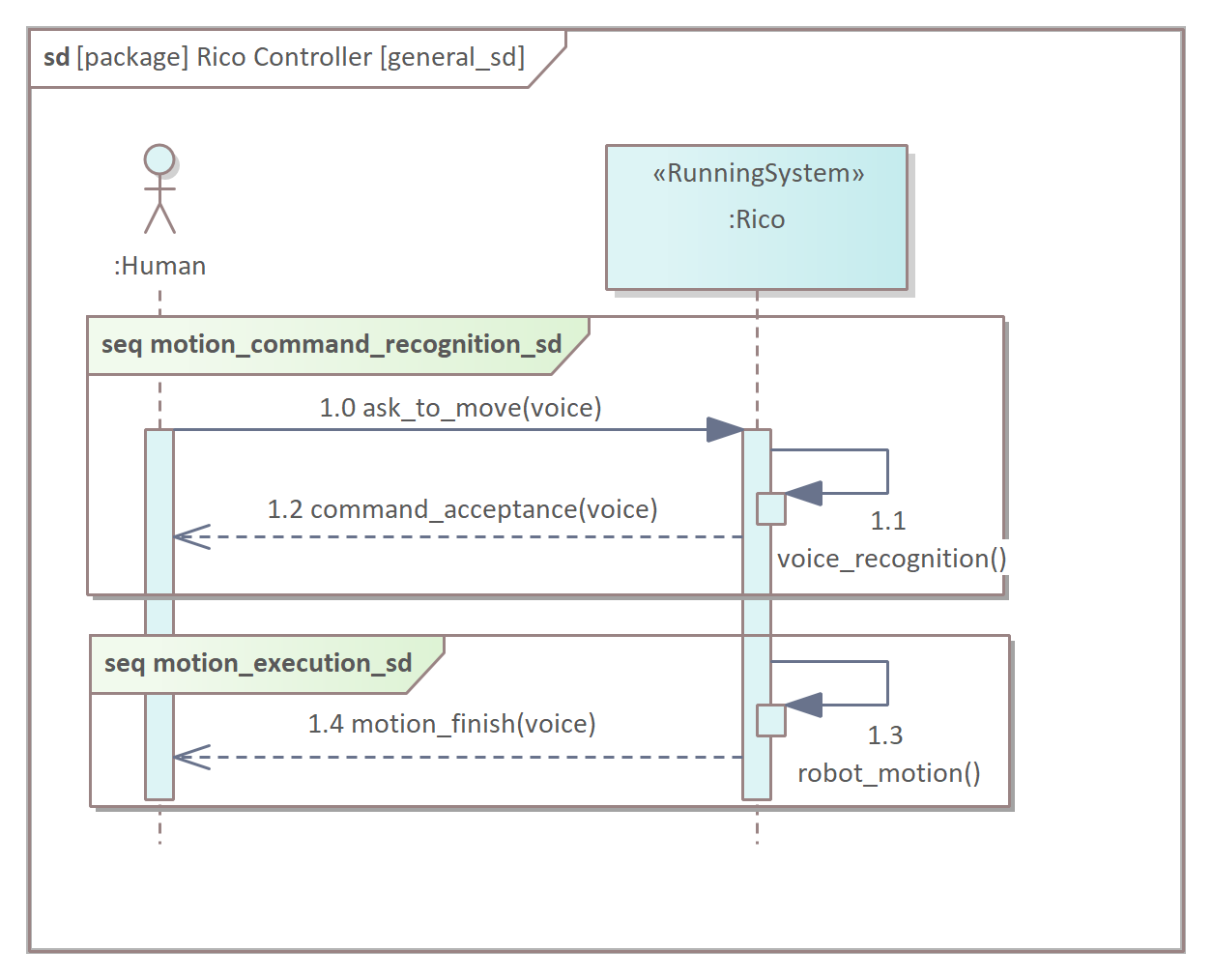}}
		\end{center}
		\caption{Concept scenario -- sd.} 
		\label{fig:general_sd}
	\end{figure}

	 The purpose of the following description is not to document the entire system but to illustrate, by example, representative aspects of the MeROS application.
		The part of the application scenario is conceptually presented in Fig.~\ref{fig:general_sd}.
	
	% DLA SKLADU ZMIENIONO WIELKOSC Z 0.7	
	
\begin{figure}[htb]
		\centering
		\begin{center}
			{\includegraphics[scale=0.7]{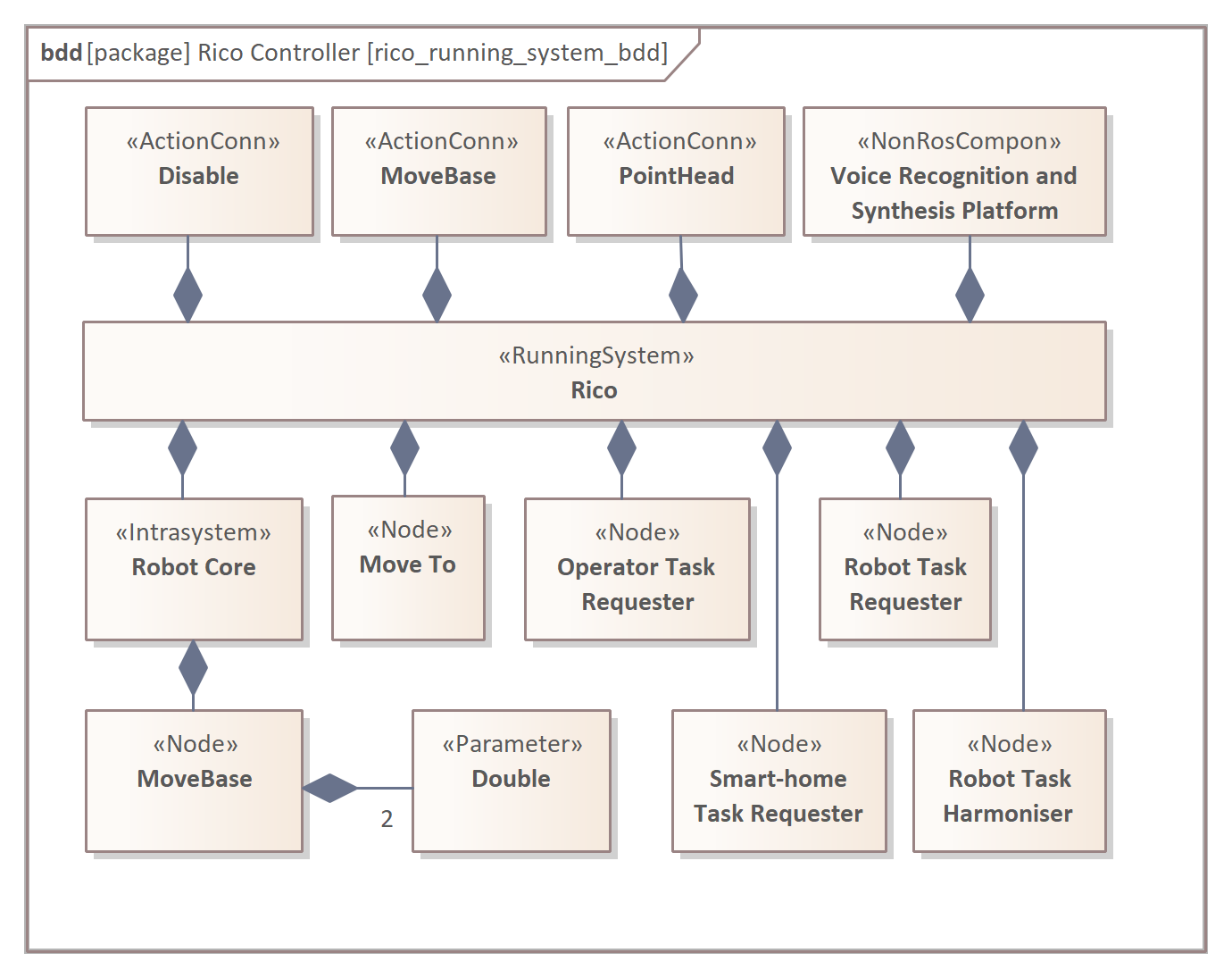}}
		\end{center}
		\caption{Rico <<RunningSystem>> composition -- bdd.}
		\label{fig:rico_running_system_bdd}
	\end{figure}

	Here, the system (<<RunningSystem>> \texttt{:Rico}) and its behaviour are formulated in a~general way. An actor (e.g. an elderly person) asks the robot to move. Then, the system recognises the voice command and vocally confirms the command's acceptance. Finally, the robot executes the motion and vocally informs that the motion is finished.
	In the following part of the description, the <<RunningSystem>> \texttt{:Rico} and sequence diagram frame \texttt{motion execution} from Fig.~\ref{fig:general_sd} are presented in a explicit way.
	The block definition diagram in Fig.~\ref{fig:rico_running_system_bdd} depicts the composition of <<RunningSystem>> \texttt{:Rico}.
	
\begin{figure}[htb]
		\centering
		\begin{center}
			{\includegraphics[scale=0.7]{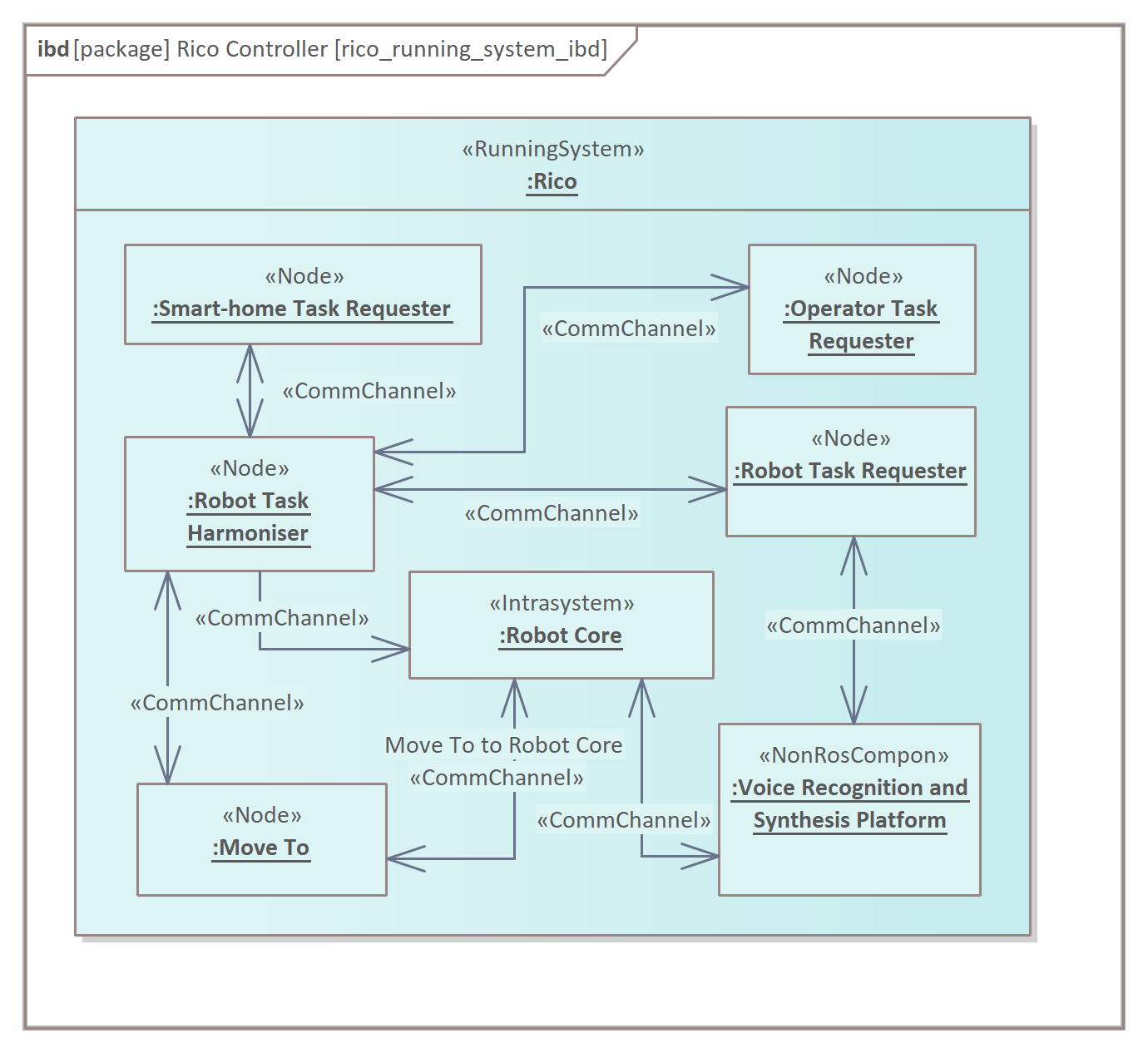}}
		\end{center}
		\caption{Structure of <<RunningSystem>> \texttt{:Rico} -- ibd.} 
		\label{fig:rico_running_system_ibd}
	\end{figure}

	The <<RunningSystem>> \texttt{:Rico} structure is depicted in Fig.~\ref{fig:rico_running_system_ibd}. Here, and in the following diagrams, the \texttt{rosout} and \texttt{ROS master} <<Node>>s were omitted to make the diagrams more compact. The specific label is needed for <<CommChannel>>, e.g., <<CommChannel>> \texttt{:Move To to Robot Core}, because this <<CommChannel>> is described later on.
	
\begin{figure}[htb]
		\centering
		\begin{center}
			{\includegraphics[scale=0.7]{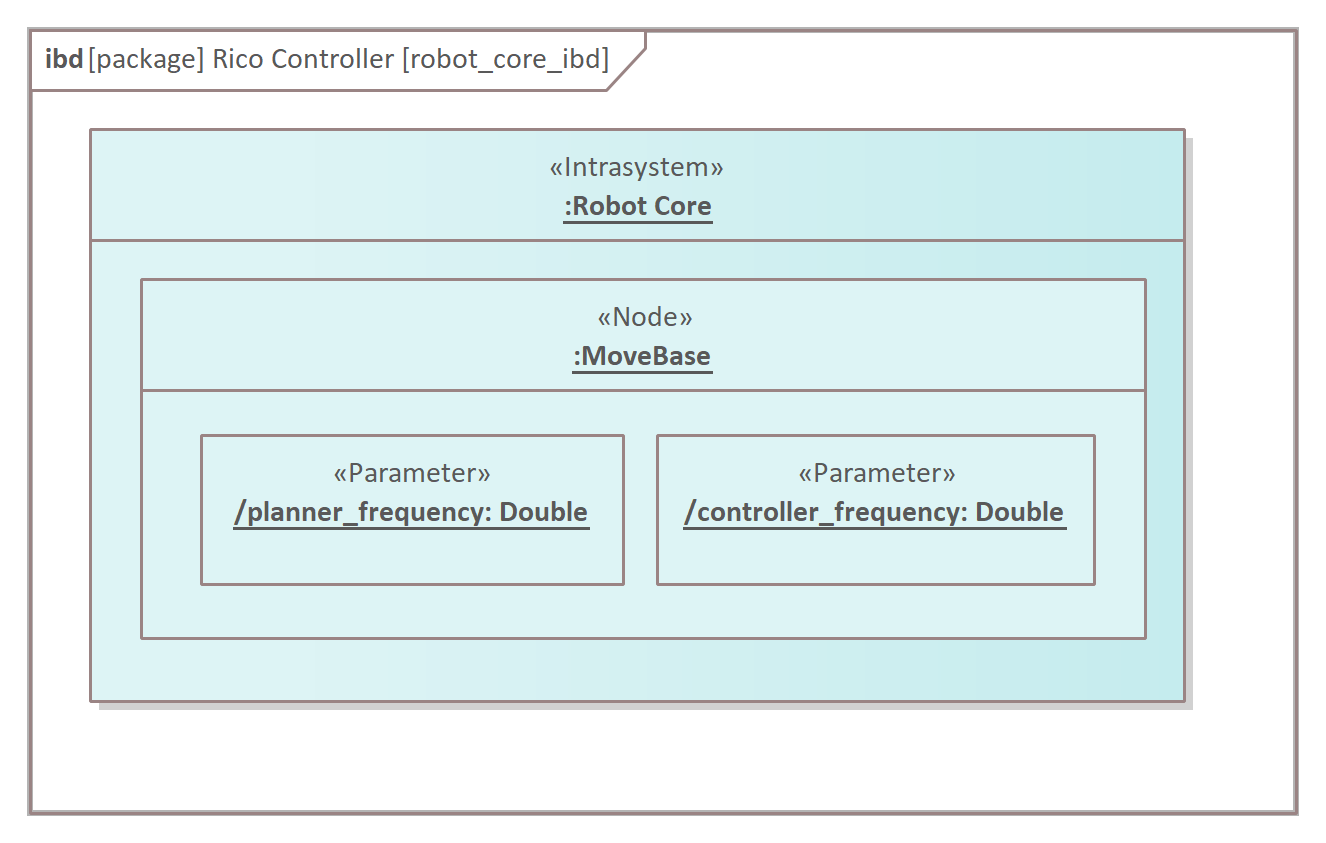}}
		\end{center}
		\caption{Selected elements of <<Intrasystem>> \texttt{:Robot Core} -- ibd.} 
		\label{fig:robot_core_ibd}
	\end{figure}

	The system is based on TaskER framework \cite{tasker2020} developed from the RAPP approach to construct systems with variable structure \cite{zielinski2017variable}. The role of the TaskER is to schedule a robot’s tasks. It consists of (i) Task Requesters <<Node>>s to submit new tasks, (ii) Task Harmoniser <<Node>> to schedule tasks execution, (iii) dynamic <<Node>>s (here, <<Node>> \texttt{:Move To}) to execute a particular task on the robot hardware and (iv) cloud part, here <<NonRosCompon>> \texttt{:Voice Recognition and Synthesis Platform}. The common part of the controller is located in <<Intrasystem>> \texttt{:Rico}.
	Fig.~\ref{fig:robot_core_ibd} illustrates how various instances of the same block are depicted in the model. Two <<Parameter>> Objects of the same classifier \texttt{:Double} are composed into <<Node>> \texttt{:MoveBase}.
	
\begin{figure}[htb]
		\centering
		\begin{center}
			{\includegraphics[width=\columnwidth]{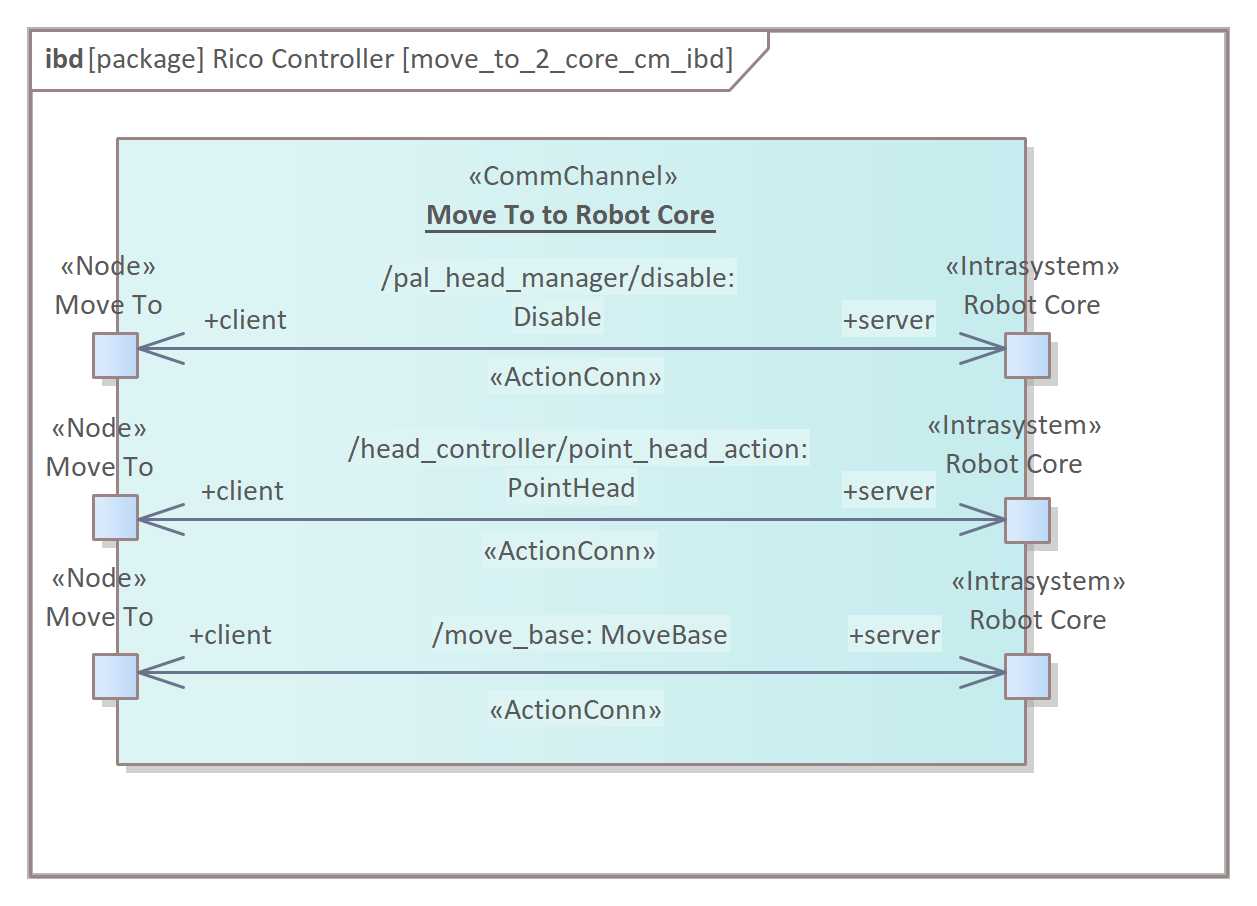}}
		\end{center}
		\caption{Example of <<CommChannel>> -- ibd.} 
		\label{fig:move_to_2_core_cm_ibd}
	\end{figure}

	<<CommChannel>> \texttt{:Move To to Robot Core} is depicted in Fig.~\ref{fig:move_to_2_core_cm_ibd}. It comprises three actions.
	
	% DLA SKLADU ZMIENIONO WIELKOSC Z 0.7	
	
\begin{figure}[htb]
		\centering
		\begin{center}
			{\includegraphics[width=\columnwidth]{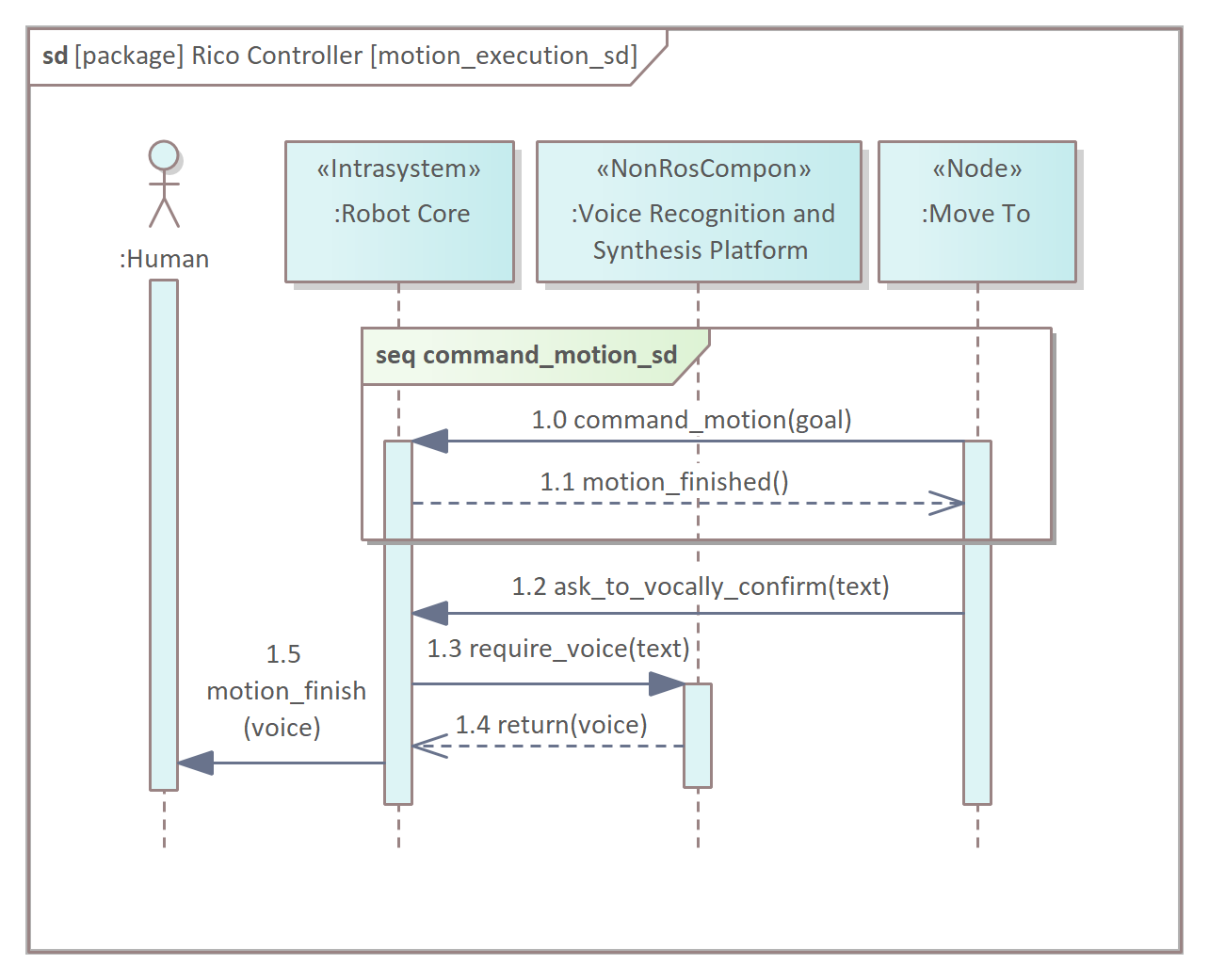}}
		\end{center}
		\caption{Motion execution operation -- sd.} 
		\label{fig:motion_execution_sd}
	\end{figure}

	The part of the scenario generally described in Fig.~\ref{fig:general_sd} is depicted in detail in Fig.~\ref{fig:motion_execution_sd}. The presentation remains conceptual from the behavioural point of view, but it considers the particular parts of the <<RunningSystem>> \texttt{:Rico}.
	
\begin{figure}[htb]
		\centering
		\begin{center}
			{\includegraphics[scale=0.7]{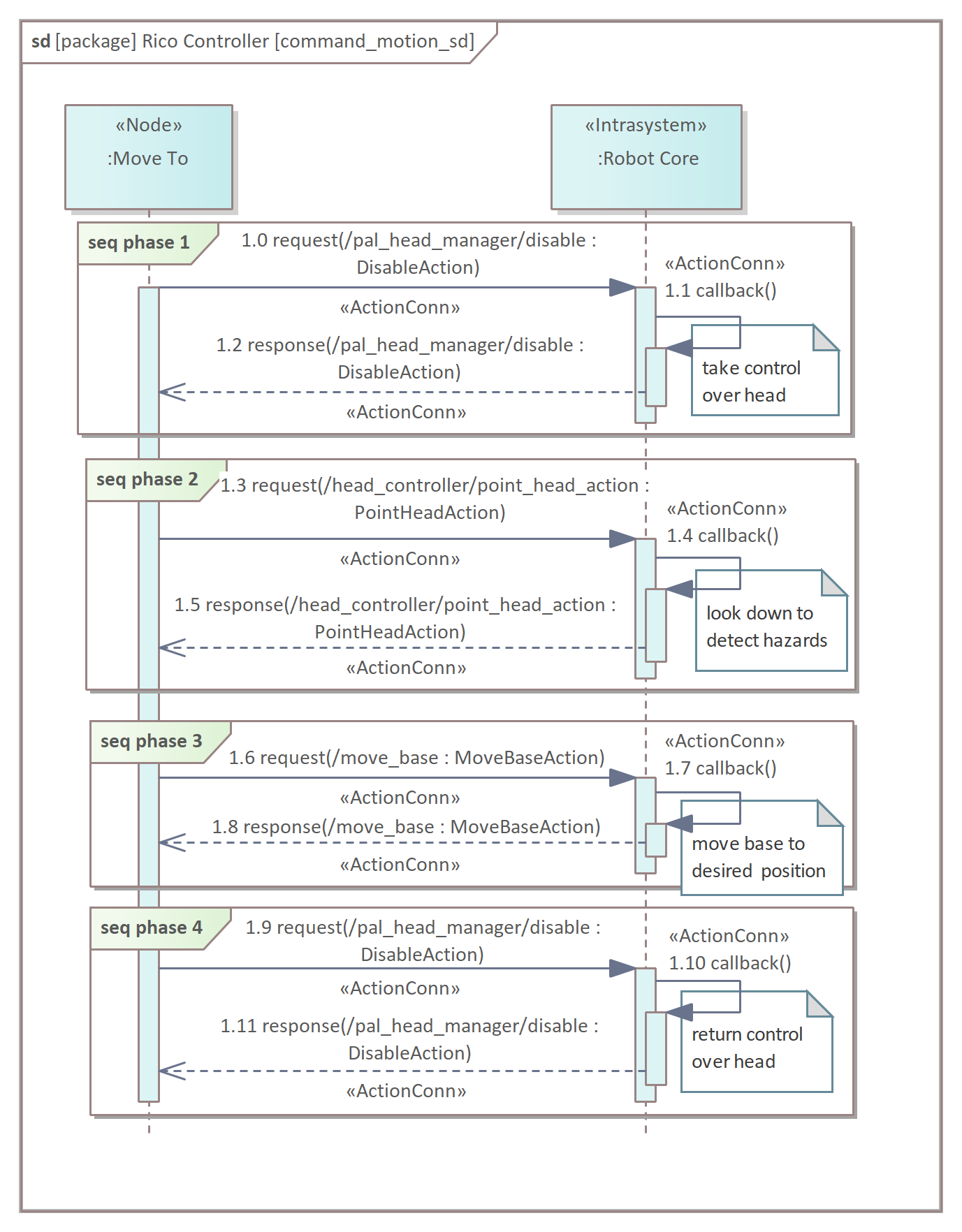}}
		\end{center}
		\caption{Command motion operation with detailed Communication methods presentation -- sd.} 
		\label{fig:command_motion_sd}
	\end{figure}

	Finally, the particular communication methods are specified on the most detailed, ROS-specific level (Fig.~\ref{fig:command_motion_sd}). The command\_motion operation includes the sequence of four steps of communication. Three Actions realise the communication, one utilised twice. The diagram comprises extra notes that make it easier to interpret.
	
\begin{figure}[htb]
		\centering
		\begin{center}
			{\includegraphics[scale=0.7]{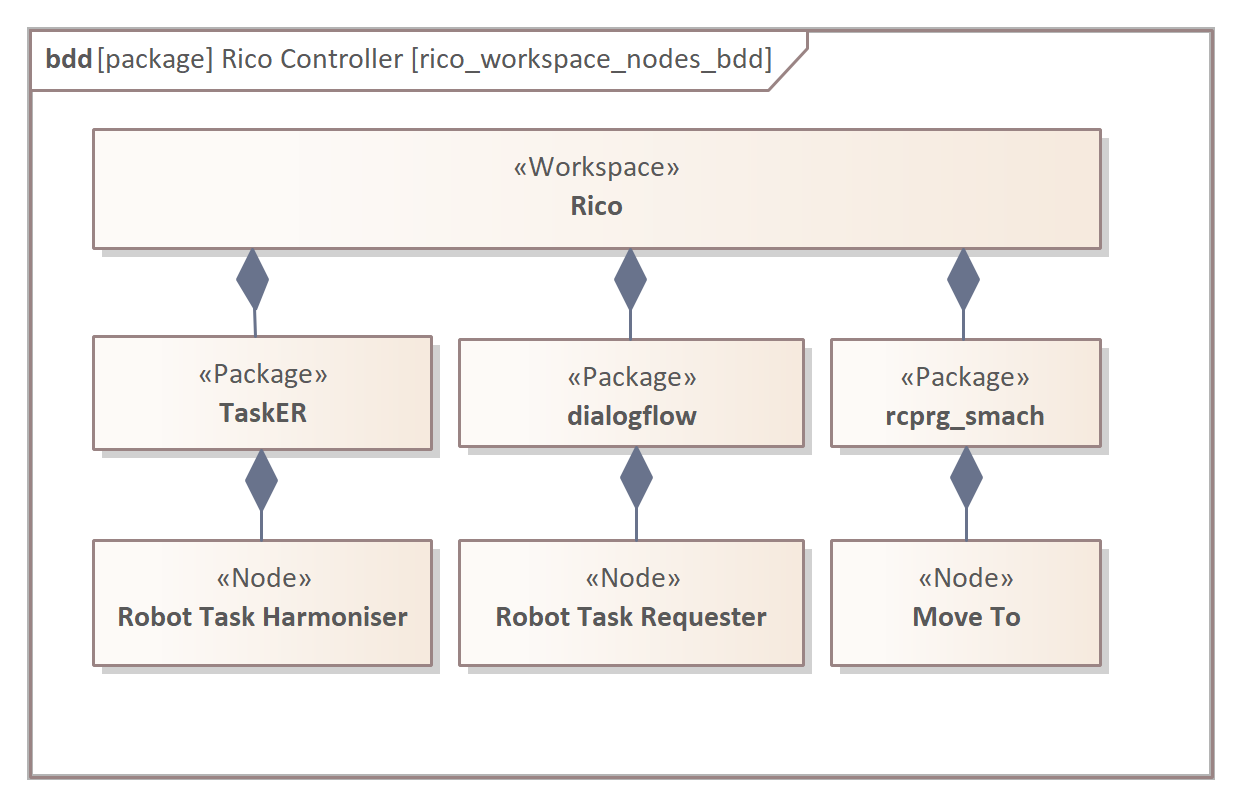}}
		\end{center}
		\caption{Rico <<Workspace>> composition -- Packages with Nodes -- bdd.}
		\label{fig:rico_workspace_nodes_bdd}
	\end{figure}

\begin{figure}[htb]
		\centering
		\begin{center}
			{\includegraphics[scale=0.7]{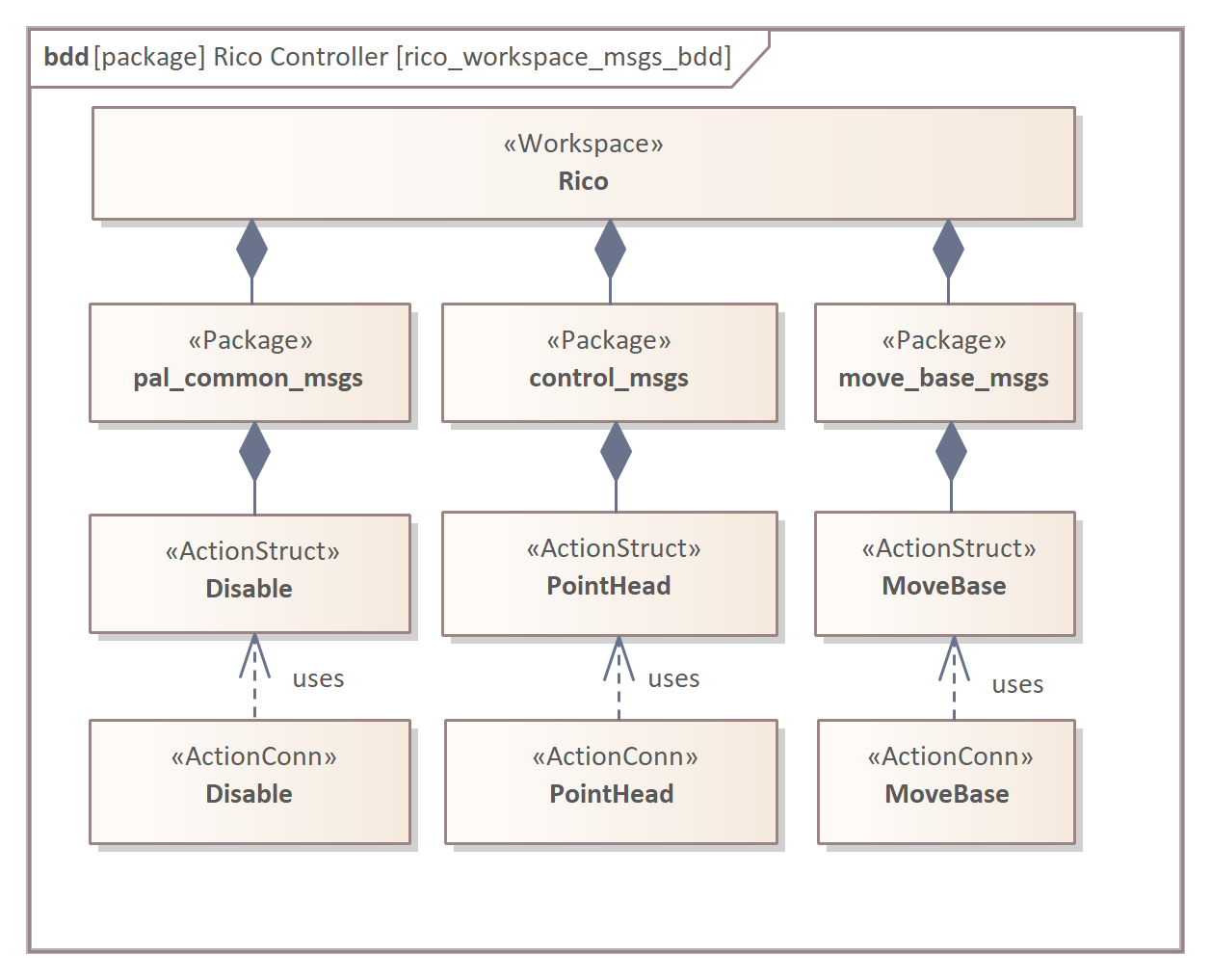}}
		\end{center}
		\caption{Rico <<Workspace>> composition -- Packages with Msgs -- bdd.}
		\label{fig:rico_workspace_msgs_bdd}
	\end{figure}

	The part of the <<Workspace>> \texttt{:Rico} that includes previously mentioned elements is presented in Fig.~\ref{fig:rico_workspace_nodes_bdd} and Fig.~\ref{fig:rico_workspace_msgs_bdd}.
	
\section{Related work}
\label{sec:related-work}

	Papers in the scope of the literature review are chosen based on an intensive study of the previous scientific work in robotic systems modelling. In particular, the survey \cite{de2021survey} is deeply analysed. As the qualification criterion for this review, the ROS metamodel was chosen that must be described using UML language or SysML language. Six papers describing five metamodels met this criterion, and all used UML to describe ROS~1. The metamodels are contrasted with the representative requirements to which MeROS is subjected (Tab.~\ref{tab:ros-spec-req-sota}) and that clearly differentiate compared metamodels.
	For clarification, the table refers to aspects of the metamodels, which are visualised in the analysed papers' diagrams.

	\begin{table}[htb]
		\centering
		\caption{MeROS requirements satisfaction in ROS specific metamodels.}
		\setlength{\tabcolsep}{3.5pt} % Default value: 6pt
		\renewcommand{\arraystretch}{1.0} % Default value: 1
		\begin{tabular}{c c c c c c c c c c c c}
			\toprule
			Metamodel/req. & \rotatebox{90}{R3.1.1 Node}  & \rotatebox{90}{R3.1.2 Nodelet}  & \rotatebox{90}{R3.1.3 ROS plugin} & \rotatebox{90}{R3.1.4 ROS library} & \rotatebox{90}{R3.2.1 Topic} & \rotatebox{90}{R3.2.2 Service} & \rotatebox{90}{R3.2.3 Action}&
			\rotatebox{90}{R3.3.1 Package} & \rotatebox{90}{R3.3.2 Metapackage}&
			\rotatebox{90}{R3.4.2 ROS Parameter} &  \rotatebox{90}{R6.3 Grouping of concepts}\\
			\midrule
			Ecore     & $+$ & $-$ & $-$ & $-$ & $+$ & $+$ & $+$ & $-$ & $-$ & $+$ & $+$\\
			RosSystem & $-$ & $-$ & $-$ & $-$ & $+$ & $+$ & $+$ & $-$ & $-$ & $-$ & $-$\\
			HyperFlex & $+$ & $-$ & $-$ & $-$ & $+$ & $+$ & $+$ & $+$ & $-$ & $+$ & $-$\\
			RoBMEX    & $+$ & $-$ & $-$ & $-$ & $+$ & $+$ & $-$ & $+$ & $+$ & $-$ & $+$\\
			ROSMOD    & $+$ & $-$ & $-$ & $+$ & $-$ & $+$ & $-$ & $+$ & $-$ & $-$ & $+$\\
			MeROS     & $+$ & $+$ & $+$ & $+$ & $+$ & $+$ & $+$ & $+$ & $+$ & $+$ & $+$\\
			\bottomrule
		\end{tabular}
		\label{tab:ros-spec-req-sota}
	\end{table}	
	
	The authors of \cite{wenger2016model} present Ecore -- the ROS 1 metamodel as the central part of the ReApp workbench created to support the efficiency of software creation for robotic systems. The metamodel is specified in a~single, extensive structural diagram, with the ROS node being its central part. The diagram describes the aspects of the running system and comprises all ROS communication methods. The nodes are integrated into an AppNetwork concept [R6.3]. 
	
	In the paper \cite{garcia2019bootstrapping}, the authors propose two methods based on the created RosSystem metamodel. It aims at the automated generation of models from manually written artefacts through static code analysis and monitoring the execution of the running system. A~large part of the work is concentrated on the toolchain. This ROS metamodel is structurally specified in a UML class diagram, emphasising communication methods.
	
	HyperFlex toolchain \cite{brugali2016hyperflex,gherardi2013variability} includes extensive and comprehensive metamodel addressing ROS 1 and complementary Orocos. Formerly, these two frameworks were used together to take from the RT properties of Orocos and the elasticity of ROS. The presentation of HyperFlex is complex \cite{brugali2016hyperflex,gherardi2013variability}, and both the running system and workspace are considered. Concepts such as nodelet or metapackage are missing due to the HyperFlex period of its foundation.
	
	RoBMEX \cite{ladeira2021robmex} was created as a~top-down methodology based on a~set of domain-specific languages that enhance the autonomy of ROS-based systems by allowing the creation of missions graphically and then generating automatically executable source codes conforming to the designed missions. Hence, the ROS metamodel was extended by the upper layer with mission/task specification. The metamodel is complex and inspiring and consists of running and workspace parts. Regarding the workspace, the grouping concepts are introduced as subpackages, classified in Tab.~\ref{tab:ros-spec-req-sota} as metapackage for generality.
	
	ROSMOD \cite{kumar2016rosmod} is the Robot Operating System Model-driven development tool suite, an integrated development environment for rapid prototyping component-based
	software for ROS. Its internal metamodel is complex and comprises a~number of standard ROS concepts and additional grouping concepts. Although the description is extensive, the ROSMOD was created in 2016, hence some current concepts are missing, like ROS actions or nodelets.

\section{Conclusions and discussion}
\label{sec:conslusions}

	Diagrams are an integral part of the description of component-based robot control systems. ROS comprises rqt\_graph tool that generates diagrams with the structure of the running system. This capability is readily used by software developers (e.g., \cite{thale2020ros,bisi2018development}) due to its ease of use. Unfortunately, this tool has many limitations despite its numerous advantages and configurability. Hence, in parallel to automatically generated diagrams, others are needed, some of which are based on UML/SysML. The most comprehensive modelling solutions include explicitly defined metamodels. As a~novelty regarding previous works, this paper proposes an up-to-date metamodel for finale release of ROS~1 and ROS~2 supported by profile to support the metamodel application in ROS applications models. In MeROS, the metamodel of original ROS concepts is extended by the abstract grouping concepts. It lets to present part of the system in a~PIM-like style instead of a~platform specific -- PSM.
		
	Although the adoption of UML/SysML-based domain metamodels has many positive implications, it also has its problems and limitations. Although the diagrams can be drawn in general-purpose graphics programs, this is not advisable, especially for complex systems. Modelling software such as Enterprise Architect or Visual Paradigm is highly recommended when creating UML/SysML projects.  In particular, creating a set of diagrams outside a SysML project is more time-consuming, and it is easier to introduce errors. In practice, the cost of modelling software is not a major obstacle, and its popularity makes it easier to implement its employment. A problem with SysML development environments is that they are not standardised in many aspects and vary considerably in functionality. This problem makes it difficult to use advanced features such as automatic model analysis, e.g. to check for metamodel compatibility.
	
	There are many methodologies for conducting and documenting projects, and not all are based on languages from the UML family. In some simplification, one could say that some project teams use UML extensively while others do not at all\footnote{\url{https://creately.com/guides/advantages-and-disadvantages-of-uml/}}. In the case of academic robotics projects, the lack of widespread UML use in earlier years may have been partly due to their typically relatively small scope. When projects are extensive and multi-asset, and the consequences of failure are high, the use of UML allows for greater efficiency of operation and reduced risk of project failure. Hence, contemporary complex robotics projects should benefit from appropriate tools to support their guidance and documentation, as has been the case for many large-scale projects, e.g., from the space industry (e.g., \cite{friedenthal2017architecting}) or the medical industry (e.g., \cite{BiomedicalHealthcare}).
	
	Many skilful and experienced programmers have not used UML\footnote{\url{https://www.techwalla.com/articles/the-disadvantages-of-uml}}. This is due to the lack of absolute necessity to use such tools both for the programming and the development of small projects. Hence, the first use of UML in a developers' team can consume a disproportionate amount of time. Another problem is the synchronisation of diagrams with source code. Here, the answer is, among other things, the appropriate level of generality of the diagrams so that unnecessary details are not mapped there. The automatic generation of code from the diagrams or the automatic generation of selected diagrams based on code can also be helpful. Finally, it is worth mentioning that UML diagrams do not constitute a complete system description. In particular, the description by diagrams can be complemented by mathematical expressions. A way of combining these two ways of description is presented in, e.g. \cite{earl2020}. SysML parametric diagrams also respond to this problem.
			
\begin{figure}[htb]
		\centering
		\begin{subfigure}[b]{.43\columnwidth}
			\includegraphics[height=5.5cm]{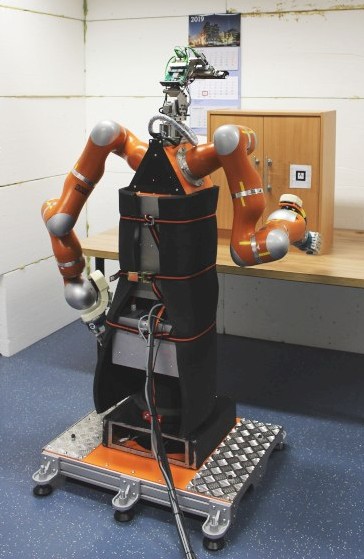}
			\caption{Velma.}
			\label{fig:velma}
		\end{subfigure}
		\begin{subfigure}[b]{.47\columnwidth}
			\includegraphics[height=5.5cm]{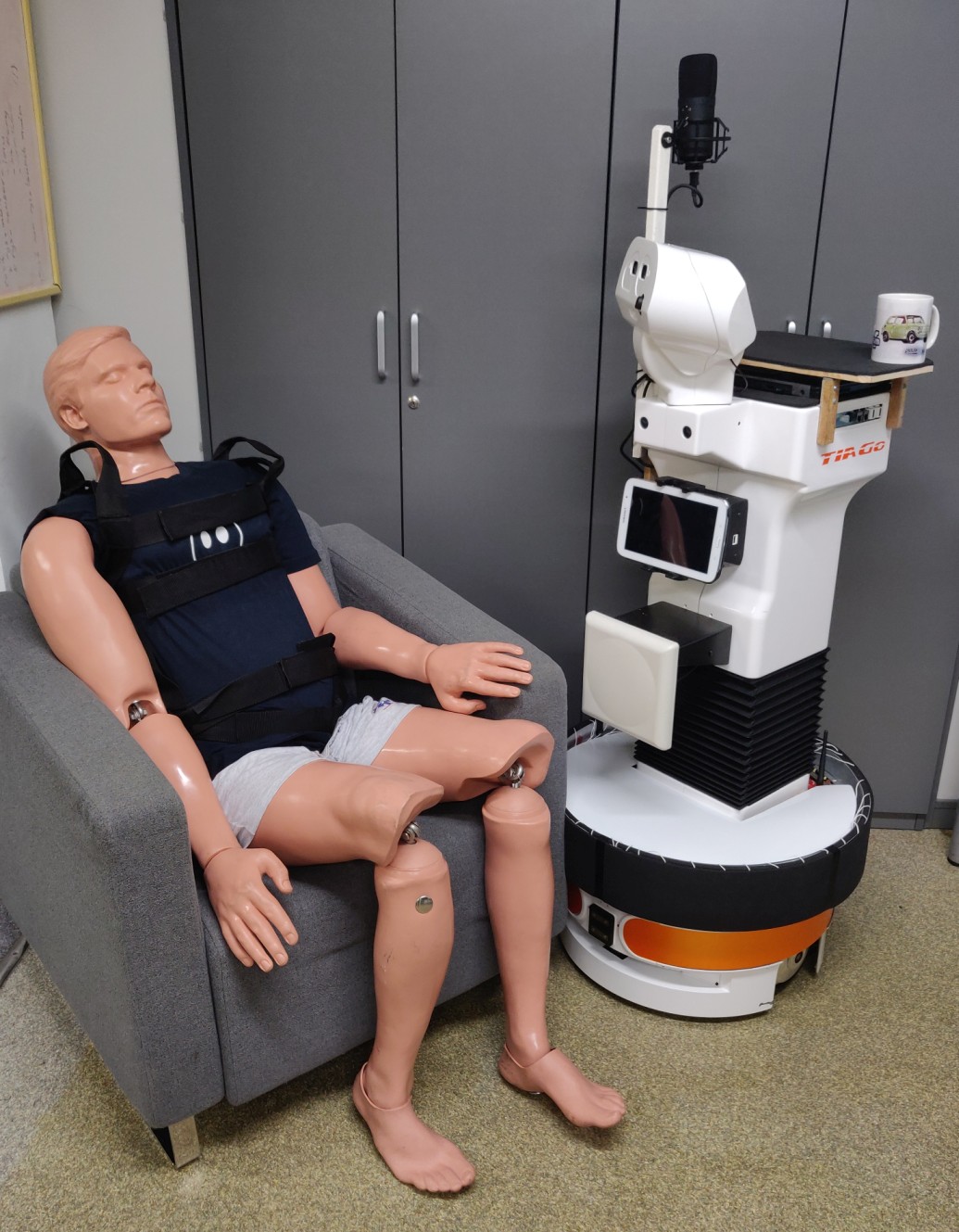}
			\caption{Rico.}\label{fig:rico}
		\end{subfigure}
		\caption{Robotic platforms specified with MeROS.}
		\label{fig:robots}
	\end{figure}

	System development involves the use of a~number of tools organised in toolchains. The degree of tools interaction varies. In software engineering, the aim is to create clear procedures for system development, indicating the dependencies between the successive stages of the development process. In robotics, there have been many works dedicated to toolchains (SmartMDSD \cite{dennis2016smartmdsd}, RobotML \cite{robotml2}, \cite{dal2022formal}), nowadays the ROS is common middleware (e.g. \cite{wienke2012meta}, BRIDE \cite{bubeck2014bride}, HyperFlex \cite{brugali2016hyperflex}). MeROS is part of the toolchain used in the Robot Programming and Machine Perception Team at Warsaw University of Technology (WUT). Currently, on the base of MeROS, the controllers are specified for the following robotic platforms:
	 \begin{itemize}
	 	\item Velma service robot  \cite{en14206693-grav-comp,Figat:2022:RAS} \footnote{\url{https://www.robotyka.ia.pw.edu.pl/robots/velma}} (Fig. \Ref{fig:velma}) (Custom controller based on ROS~1 and Orocos for hardware control and simulation in Gazebo),
	 	\item assistive robot Rico \cite{tasker2020,karwowski2021hubero} \footnote{\url{https://www.robotyka.ia.pw.edu.pl/robots/rico}} (Fig.\ref{fig:rico}) (Extended PAL controller based on ROS~1 and Orocos, recent works with ROS~2 in simulation),
	 	\item MiniRyś -- mobile robot with various modes of locomotions \footnote{\url{https://www.robotyka.ia.pw.edu.pl/robots/minirys}} (Custom controller based on ROS~2 for hardware control and simulation in Gazebo),
	 	\item Dobot Magician -- portable, 4-DOF robotic manipulator \footnote{\url{https://www.robotyka.ia.pw.edu.pl/robots/magician}} (Custom controller based on ROS~2 for hardware control).
	 \end{itemize}

	% The height can be increased to 5.5cm
		
	At the forefront of the toolchain stays the modelling of the system with PIM using the EARL-based \cite{earl2020} SPSysML \cite{dudek2023spsysml}. EARL is derived from agent theory \cite{kornuta-bpan-2020, zielinski2010motion,zielinski2017variable}. Nowadays, in the intermediate stage, MeROS plays the major role as a~PSM. Finally, FABRIC  \cite{Seredynski-fabric-romoco-2019}, as well as alternative approaches \cite{winiarskimmar2015,figat2020robotic} are used to support code generation. Current work concerns deepening the integration of MeROS with the rest of the toolchain.

	\bibliography{meros}

	\begin{IEEEbiography}[{\includegraphics[width=1in,height=1.25in,clip,keepaspectratio]{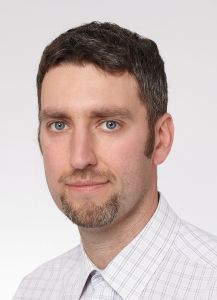}}]{Tomasz Winiarski} IEEE, INCOSE Member, M.Sc./Eng. (2002), PhD (2009) in control and robotics, from Warsaw University of Technology (WUT), assistant professor of WUT. He is a~member of the Robotics Group as the head of the Robotics Laboratory in the Institute of Control and Computation Engineering (ICCE), Faculty of Electronics and Information Technology (FEIT). He is working on the modelling and design of robots and programming methods of robot control systems. The research targets service and social robots as well as didactic robotic platforms. His personal experience concerns the development and modelling of robotic frameworks, manipulator position--force and impedance control, safety in robotic research. Recently, he was the head of the WUT group in AAL -- INCARE project "Integrated Solution for Innovative Elderly Care".
	\end{IEEEbiography}

	\EOD
	
\end{document}